\documentclass{article} 
\usepackage{iclr2025,times}

\usepackage{amsmath,amsfonts,bm}

\def\eqref#1{equation~\ref{#1}}

\def\1{\bm{1}}

\DeclareMathAlphabet{\mathsfit}{\encodingdefault}{\sfdefault}{m}{sl}
\SetMathAlphabet{\mathsfit}{bold}{\encodingdefault}{\sfdefault}{bx}{n}

\usepackage{amssymb}
\usepackage{pifont}
\usepackage{bibentry}
\usepackage[utf8]{inputenc} 
\usepackage[T1]{fontenc}    
\usepackage[pagebackref=true,breaklinks=true,letterpaper=true,colorlinks,bookmarks=false]{hyperref}
\hypersetup{citecolor=[RGB]{119,185,0}}
\usepackage{times}
\usepackage{latexsym}
\usepackage{xcolor}
\usepackage{makecell}
\usepackage{times}
\usepackage{hyperref}
\usepackage{inconsolata}
\usepackage{url}
\usepackage{multirow}
\usepackage[pdftex]{graphicx}
\usepackage{caption}
\usepackage{subcaption}
\usepackage{amsmath}
\usepackage{bbm}
\usepackage{booktabs}
\usepackage{algorithmic}
\usepackage{amssymb}
\usepackage{bm}
\usepackage[ruled,vlined,linesnumbered]{algorithm2e}
\usepackage{enumitem}
\usepackage{pifont}
\usepackage{pbox}
\usepackage[utf8]{inputenc} 
\usepackage[T1]{fontenc}    
\usepackage{url}            
\usepackage{booktabs}       
\usepackage{amsfonts}       
\usepackage{nicefrac}       
\usepackage{microtype}      
\usepackage{xcolor}         
\usepackage{tabularx}
\usepackage{url}            
\usepackage{booktabs}       
\usepackage{amsfonts}       
\usepackage{nicefrac}       
\usepackage{microtype}      
\usepackage{graphicx}
\usepackage{enumitem}
\usepackage{multirow}
\usepackage{epsfig}
\usepackage{wrapfig}
\usepackage{array}
\usepackage{makecell}
\usepackage{textcomp,booktabs}
\usepackage{colortbl}
\usepackage{amsmath}
\usepackage{xcolor}
\definecolor{mygray}{gray}{.9}

\title{Eagle: Exploring The Design Space for Multi-\\modal LLMs with Mixture of Encoders}

\author{Min Shi$^{2}$\thanks{Equal contribution. Work done during an internship at NVIDIA.}~\,,
Fuxiao Liu$^{3*}$,
Shihao Wang$^{4}$,
Shijia Liao$^1$,
Subhashree Radhakrishnan$^1$,\\
\textbf{Yilin Zhao}$^5$,
\textbf{De-An Huang}$^1$,
\textbf{Hongxu Yin}$^1$, 
\textbf{Karan Sapra}$^1$, 
\textbf{Yaser Yacoob}$^3$, 
\textbf{Humphrey Shi}$^2$,\\
\textbf{Bryan Catanzaro}$^1$,
\textbf{Andrew Tao}$^1$, 
\textbf{Jan Kautz}$^1$, 
\textbf{Zhiding Yu}$^1$\thanks{Equal advising. Corresponding authors: \url{{guilinl, zhidingy}@nvidia.com}.}~\,,
\textbf{Guilin Liu}$^{1\dagger}$
\\[0.25cm]
$^1$NVIDIA~~~~$^2$Georgia Tech~~~~$^3$UMD~~~~$^4$HKPU~~~~$^5$NYU\\
\href{https://github.com/NVlabs/Eagle}{https://github.com/NVlabs/Eagle}
}

\iclrfinalcopy 
\begin{document}

\maketitle
\begin{figure*}[ht]
    \centering\vspace{-0.2in}
    \includegraphics[width=\linewidth]{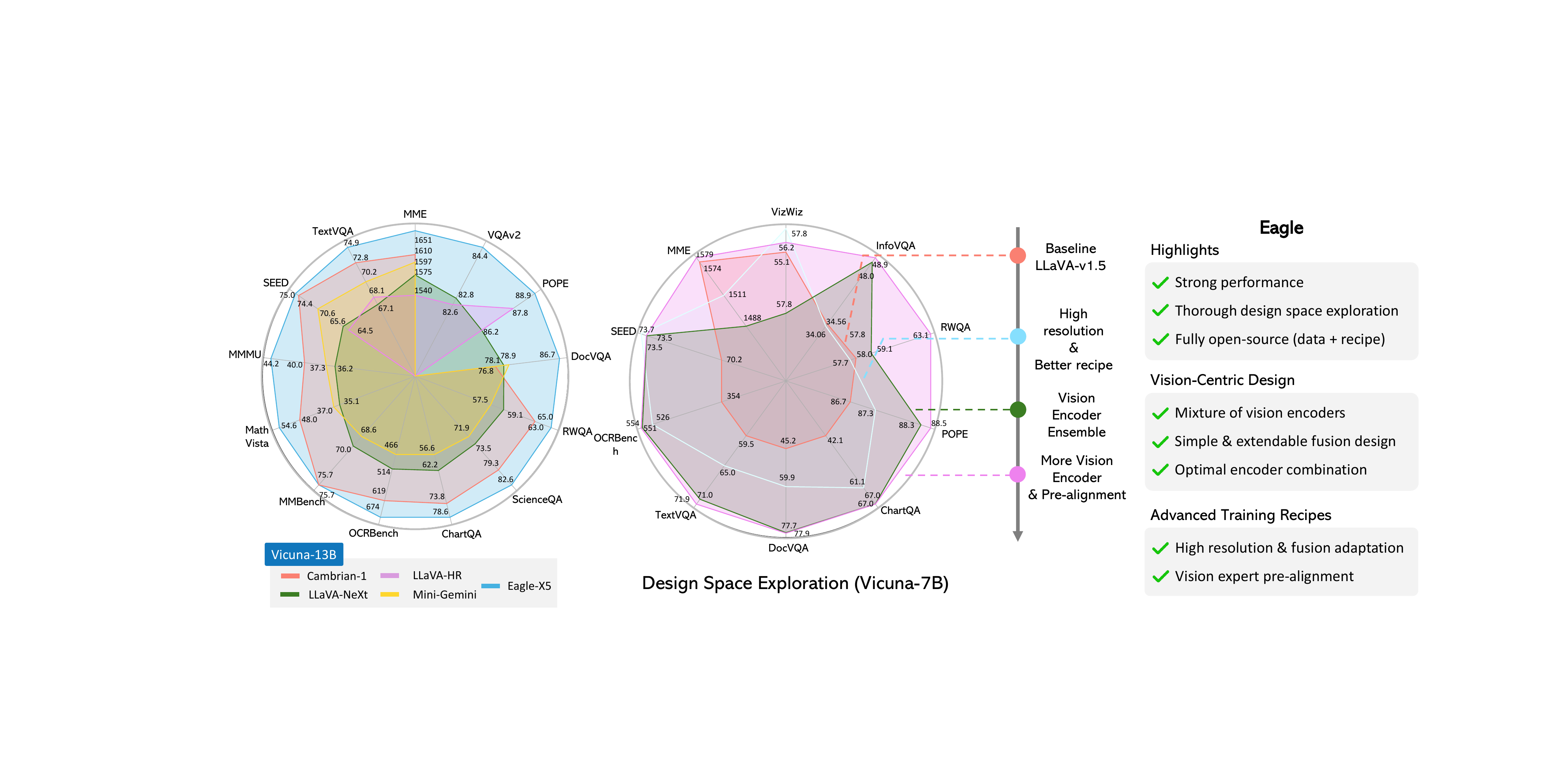}
    \caption{ \textbf{Overview of \textit{Eagle}.} \textit{Eagle} is a family of multimodal large language models (MLLMs) with a mixture of vision encoders. Left: comparisons between \textit{Eagle} and existing competitive MLLMs with \textit{Vicuna-13B}~\citep{vicuna2023}, with \textit{Eagle} achieving favorable results on all 13 benchmarks. Middle: an evolutionary road map of the design space and advanced training recipes leading to consistent and significant improvements. Right: highlights and core features of \textit{Eagle}.
    }
    \label{fig:Teaser_figure}
\end{figure*}

\maketitle

\begin{abstract}
The ability to accurately interpret complex visual information is a crucial topic of multimodal large language models (MLLMs). Recent work indicates that enhanced visual perception significantly reduces hallucinations and improves performance on resolution-sensitive tasks, such as optical character recognition and document analysis. A number of recent MLLMs achieve this goal using a mixture of vision encoders. Despite their success, there is a lack of systematic comparisons and detailed ablation studies addressing critical aspects, such as expert selection and the integration of multiple vision experts. This study provides an extensive exploration of the design space for MLLMs using a mixture of vision encoders and resolutions. Our findings reveal several underlying principles common to various existing strategies, leading to a streamlined yet effective design approach. We discover that simply concatenating visual tokens from a set of complementary vision encoders is as effective as more complex mixing architectures or strategies. We additionally introduce \textit{Pre-Alignment} to bridge the gap between vision-focused encoders and language tokens, enhancing model coherence. The resulting family of MLLMs, \textit{Eagle}, surpasses other leading open-source models on major MLLM benchmarks.
\end{abstract}

\vspace{-0.05in}
\section{Introduction}
\vspace{-0.05in}
The success of large language models (LLMs) has triggered significant interest in enabling their visual perception capability, such that they could see, understand, and reason in the real world. At the center of these multimodal large language models (MLLMs)~\citep{fei2024multimodal} is a typical design where images are converted into a series of visual tokens by the vision encoders and appended with the text embeddings. \textit{CLIP}~\citep{radford2021learning} is often chosen as the vision encoder since its visual representation is aligned with the text space by pre-training on image-text pairs. Depending on the architectures, training recipes, and the way how vision tokens are injected into the language model, there exist various notable families of MLLMs such as \textit{Flamingo}~\citep{alayrac2022flamingo}, \textit{BLIP}~\citep{li2022blip,li2023blip,dai2024instructblip}, \textit{PaLI}~\citep{chen2023pali}, \textit{PaLM-E}~\citep{driess2023palm} and \textit{LLaVA}~\citep{liu2023visual,liu2023improved}. Most of these works keep relatively low input resolutions due to the limits on pre-trained vision encoders and LLM sequence length.

Recent studies~\citep{li2024monkey,liu2024llavanext} show that stronger vision encoder design is important for mitigating MLLM hallucinations~\citep{liu2023hallusionbench,wu2024safety} and improving resolution-sensitive tasks like optical character recognition (OCR). A constellation of works thus focuses on enhancing the capability of the vision encoder. For example, scaling up the pre-training data and parameters of vision encoder~\citep{chen2023internvl} or dividing images into low-resolution patches~\citep{liu2024llavanext,shi2024need}. However, these approaches usually introduce large training resources. An efficient yet powerful strategy is to mix visual encoders pre-trained with different tasks and input resolutions, either fusing higher resolution encoders with the \textit{CLIP} encoder~\citep{luo2024feast,li2024mini}, sequentially appending features from different encoders~\citep{fan2024mousi,lin2023sphinx,karamcheti2024prismatic,tong2024cambrian}, or adopting more complex fusion and routing strategies to make the best of different encoders~\citep{lee2024moai,zong2024mova}. In addition, Prismatic VLM~\citep{liu2024prismer} incorporates multiple vision encoders with channel concatenation as part of their design space exploration. These ``mixture-of-vision-experts'' strategies are shown to be effective. However, a detailed study focusing on their designs is still lacking.

To address the above questions, our work systematically investigates the mixture-of-vision-encoders design space for improved MLLM perception. As shown in Fig.~\ref{fig:Teaser_figure}, our exploration of the design space consists of the following steps: 1) Benchmarking various vision encoders and searching recipes for higher resolution adaptation; 2) ``Apples to apples'' comparison between vision encoder fusion strategies; 3) Progressive identification of the optimal combination of multiple vision encoders; 4) Improved vision expert pre-alignment and data mixture. Our study covers the performance of vision encoders pre-trained on different tasks and resolutions (e.g., vision-language alignment~\citep{ilharco_gabriel_2021_5143773,cherti2023reproducible,radford2021learning,schuhmann2022laionb}, self-supervised learning~\citep{oquab2023dinov2}, detection~\citep{fang2023eva,fang2023eva02}, segmentation~\citep{kirillov2023segment}, and OCR~\citep{lee2023pix2struct}). We use a round-robin approach to incorporate additional vision experts. Starting with the basic \textit{CLIP}~\citep{radford2021learning} encoder, we add one additional expert each time with the best improvement in each round.

Our work is not the first one to leverage multiple vision encoders in MLLM. However, the systematic study leads to several interesting new findings under this setting:

\begin{itemize}[leftmargin=6mm]
    \item Unlocking the vision encoders during MLLM training matters. This is in sharp contrast to the LLaVA~\citep{liu2023visual,liu2023improved} family and many works that consider multiple vision encoders or teachers~\citep{lin2023sphinx,liu2024prismer,fan2024mousi,kar2024brave,radio,lee2024moai}, where freezing the vision encoders has been a common choice.
    \item Some recently proposed fusion strategies~\citep{luo2024feast,li2024mini} do not show significant advantages despite their advanced designs. Instead, we find that straightforward channel concatenation stands out as a simple yet competitive fusion strategy, offering the best efficiency and performance.
    \item Incorporating additional vision experts leads to consistent gain, making it a promising path to systematically enhance MLLM perception besides scaling up vision encoders. The improvement is particularly pronounced when vision encoders are unlocked.
    \item We propose a pre-alignment stage where non-text-aligned vision experts are individually fine-tuned with a frozen LLM before being trained together. This stage is found to enhance the MLLM performance significantly under the mixture-of-vision-encoder design.
\end{itemize}

We finally conclude our findings into a family of MLLMs termed \textit{Eagle}. \textit{Eagle} is evaluated on a series of benchmarks, including visual question answering, OCR/document-related tasks, and benchmarks tailored for MLLMs. Our model attains state-of-the-art performance across different benchmarks and demonstrates obvious advantages on OCR and document understanding tasks. Using the same pre-train and supervised fine-tuning data from \textit{Cambrian-1}~\citep{tong2024cambrian} - a concurrent family of vision-centric MLLMs sharing similar design spirits, \textit{Eagle} models overall achieve better performance. We hope that the \textit{Eagle} can provide a highly performant and easy-to-reproduce MLLM solution to the community.

\vspace{-0.05in}
\section{Design space exploration}
\label{sec:preliminary}
\vspace{-0.05in}

\begin{figure*}[t]
    \centering
    \includegraphics[width=1.0\textwidth]{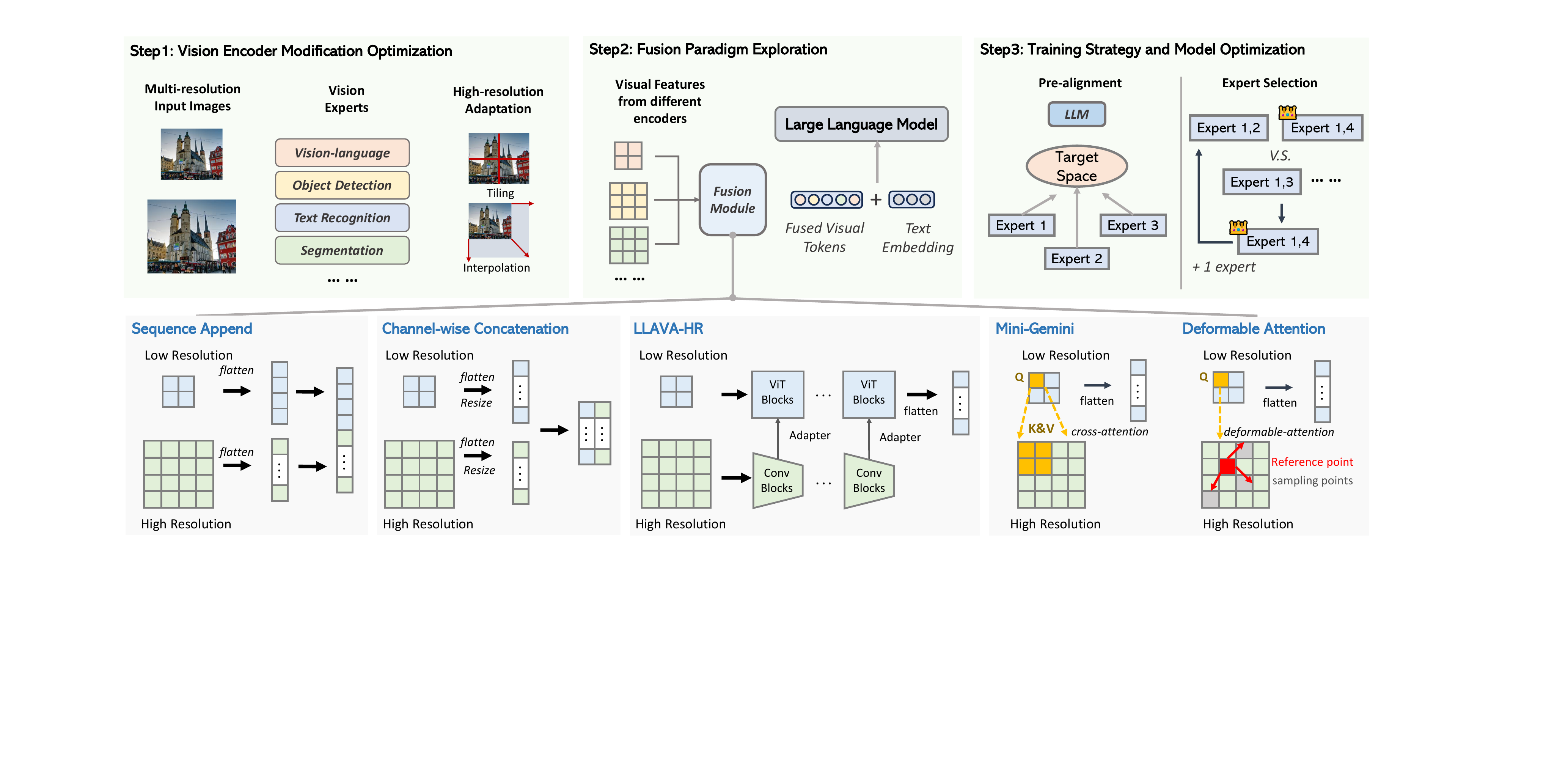}
    \vspace{-2mm}
    \caption{
    \textbf{Overview of the \textit{Eagle} exploration}. In this work, we explore the design space of Multi-Modal Large Language Models (MLLMs) with multiple vision encoders, aiming to identify optimized design choices and enhance MLLM perception. We collect a range of vision experts and adapt them for integration into MLLMs. A systematic comparison of popular fusion paradigms is then conducted under controlled settings. After identifying discrepancies between vision experts pre-trained on different tasks, we optimize the pre-training strategy through a pre-alignment stage and use round-robin searching to determine the optimal combination of vision encoders.
    }\label{fig:workflow_figure}
    \vspace{-0.1in}
\end{figure*}

In this section, we show how to utilize the advantages of different vision encoders via step-by-step investigations, yielding the \textit{Eagle} model family. Unlike previous methods focusing on new fusion strategies or architectures among vision encodes, our goal is to identify a set of minimalistic design to fuse different vision encoders supported with detailed ablations, removing any unnecessary parts. As shown in Fig.~\ref{fig:workflow_figure}, we start by extending the basic \textit{CLIP} encoder~\citep{radford2021learning} to a set of vision experts with different architectures, pre-training tasks, and resolutions. With these experts, we then compare different fusion architectures and methods and study how to optimize the pre-training strategies given more encoders. We also give a detailed analysis of how to select the vision encoders to be integrated. Finally, we put all the findings together and further extend to multiple expert vision encoders with different resolutions and domain knowledge. 

\vspace{-0.05in}
\subsection{Base setup}
\vspace{-0.05in}
\label{sec:baseline-setup}
We adopt \textit{LLaVA-1.5}'s~\citep{liu2023improved} model architecture as the basis, which consists of a large language model (Vicuna-v1.5 7B~\citep{vicuna2023}), a vision encoder, and a projection layer. The projection layer projects the visual embedding from the vision encoder into the text embedding space.

\noindent \textbf{Base training data.}
We adopt the same pre-training data (\textbf{LLaVA-595k}) as \textit{LLaVA-1.5}~\citep{liu2023improved} for the first pre-training stage, which consists of \textit{$595$k} image text pairs. To fully examine the potential of different vision experts and fusion methods, instead of using the SFT data from \textit{LLaVA-1.5}~\citep{liu2023improved}, we collect data from a series of tasks and convert them into multimodal conversations for the supervised fine-tuning (SFT) stage, denoted as \textbf{Eagle1.8M} in Table~\ref{tab:sft_base}.

\vspace{-1mm}
\begin{table}[h]
    \caption{Composition of the base supervised fine-tuning data (Eagle1.8M).}\vspace{-2mm}
    \label{tab:sft_base}
    \small
    \centering
    \renewcommand{\arraystretch}{1.2} 
    \resizebox{\textwidth}{!}{
    \addtolength{\tabcolsep}{-4pt}
    \begin{tabular}{c|c}
    \toprule
    Total Data Size & Data Source \\ 
    \midrule
    1,809k & \makecell[c]{\textit{LLaVA}-1.5 (665k)~\citep{liu2023improved}, DocVQA (39k)~\citep{mathew2021docvqa}, synDog-EN (50k)~\citep{kim2022donut},\\
    ChartQA (28k)~\citep{masry2022chartqa}, DVQA (25k)~\citep{kafle2018dvqa}, AI2D (15k)~\citep{Kembhavi2016ADI},\\
    ShareGPT-4V (100k)~\citep{chen2023sharegpt4v}, laion-GPT4V (11k)~\citep{laion-gpt4v}, LVIS-Instruct4V (220k)~\citep{wang2023instruct4v},\\
    LRV-Instruct (150k)~\citep{liu2023aligning}, Geo170k (120k)~\citep{geo170k}, LLaVAR (20k)~\citep{zhang2023llavar},\\ 
    Visual7W (70k)~\citep{zhu2016cvpr}, Open-Hermes 2.5 (300k)~\citep{OpenHermes2.5}}\\
    \bottomrule
    \end{tabular}
    }
\end{table}
\vspace{-0.05in}

\noindent \textbf{Base training recipes.}
We start from the \textit{LLaVA-1.5}~\citep{liu2023improved} recipe where the model is first pre-trained with image-text pairs for one epoch with a batch size of 256. The whole model is frozen and only the projector layer is updated in this pre-training stage. In the second stage, we further fine-tune the model on the multi-modal conversation data for one epoch with a batch size of 128. The learning rates are set to be \textit{1e-3} for the first stage and \textit{2e-5} for the second stage, respectively.

\noindent \textbf{Base evaluation.}
To conduct a comprehensive comparison of various methods, we adopt 11 distinct benchmarks that span multiple tasks. These benchmarks include 
1) \textbf{General VQA tasks}: GQA~\citep{hudson2019gqa}, VizWiz~\citep{2018vizwiz}, MME~\citep{fu2023mme}, SEED~\citep{li2023seed};
2) \textbf{OCR/document/chart understanding}: OCRBench~\citep{liu2023hidden}, 
DocVQA~\citep{mathew2021docvqa}, 
ChartQA~\citep{masry2022chartqa};  
3) \textbf{vision-centric tasks}: POPE~\citep{li2023evaluating}, RealWorldQA~\citep{xai2024grok}; 
4) \textbf{knowledge-based tasks}:
ScienceQA~\citep{saikh2022scienceqa}, 
AI2D~\citep{ai2d}. 
To obtain an average score, we normalize each benchmark to a total score of 1,000 and then calculate the average score across all benchmarks.

\vspace{-0.05in}
\subsection{Stronger \textit{CLIP} encoder}
\label{sec:hr-adaption}
\vspace{-0.05in}

We start our exploration by upgrading the vanilla \textit{CLIP}~\cite{radford2021learning} model since it has become a standard choice for most of the MLLMs~\cite{liu2023visual,liu2023improved}. While \textit{CLIP} models are known to benefit multimodal tasks via the text-image alignment, they also have inherent drawbacks. For instance, many existing MLLMs~\citep{liu2023improved} tend to use the pre-trained \textit{CLIP} resolutions (such as $224\times 224$ or $336\times 336$) as their input resolutions. In these cases, the encoders often fail to capture fine-grained details that are important for resolution-sensitive tasks like OCR and document understanding~\citep{li2024monkey}. 

To handle increased input resolution, a common practice is to use tiling where input images are divided into tiles and encoded separately~\citep{liu2024llavanext,li2024monkey}, or just interpolate the position embedding of the vision transformer model to fit high-resolution inputs~\citep{chen2023palix,chen2023pali3,beyer2024paligemma}. We compare these two approaches with frozen/unfrozen vision encoders under different resolutions, with the results shown in Table~\ref{tab:resolution}. Our findings can be summarized as follows:

\begin{itemize}[leftmargin=6mm]
    \item Updating the \textit{CLIP} encoder during SFT significantly improves performance at higher resolutions but slightly reduces it when using the pre-training resolution.
    \item Interpolating CLIP encoder to fit the input size of $448 \times 448$ offers a strong balance between efficiency and performance, trailing the $672 \times 672$ version with less than half the tokens.
    \item Despite its smaller size (0.3B vs. 5.9B) and less pre-training data, the \textit{CLIP} encoder gets close with interpolation approaches \textit{InternVL}'s~\citep{chen2023internvl} performance under the same setting.
\end{itemize}

\begin{wrapfigure}{rt}{0.5\textwidth}
    \vspace{-12pt}
    \makeatletter\def\@captype{table}\makeatother 
    \caption{\textbf{Comparing different high-resolution adaption methods.} ``\#Tok(V)'' denotes the number of visual tokens. ``\#Parames'', ``FLOPs'' and ``Img/Sec'' denote the model size, complexity and throughput (bs=4) of the vision encoder.
    }\vspace{-4pt}
    \label{tab:resolution}
    \small
    \centering
    \renewcommand{\arraystretch}{1.2} 
    \addtolength{\tabcolsep}{-4pt}
    \resizebox{\linewidth}{!}{
    \begin{tabular}{ccccccccccccccccccc}
    \toprule[1.5pt]
    Method & Unfreeze & Res. &\#Tok(V) & \#Params & FLOPs & Img/Sec & Avg. \\
    \midrule
    \textit{Original} & \ding{55} & 336 & 576 & 0.3B & 119G & 197.2 & 616.5 \\
    \textit{Original } & \ding{51} & 336 & 576 & 0.3B & 119G & 197.2 & 562.6 \\
    \midrule
    \textit{Interpolate} & \ding{55} & 448 & 1024 & 0.3B & 214G & 119.5 & 589.7 \\
    \textit{Interpolate} & \ding{51} & 448 & 1024 & 0.3B & 214G & 119.5 & 670.5 \\
    \textit{Interpolate} & \ding{51} & 672 & 2304 & 0.3B & 480G & 56.3 & \textbf{674.2} \\
    \textit{Tiled-input} & \ding{51} & 672 & 2304 & 0.3B & 476G & 51.6 & 673.9 \\
    \midrule
    \midrule
    \textit{InternVL}    & \ding{55} & 448 & 1024 & 5.9B & 5669G & 13.52 & 661.9 \\
    \textit{InternVL}    & \ding{51} & 448 & 1024 & 5.9B & 5669G & 13.52 & 671.5 \\
    \bottomrule[1.5pt]
    \end{tabular}
    }
    \vspace{-24pt}
\end{wrapfigure}

Based on the results, we can see that \textit{Direct interpolation} to $448\times 448$ can achieve competitive performance while being more efficient. We thus use the \textit{CLIP} encoder with a $448 \times 448$ input resolution while unlocking the encoder during the SFT stage.

\subsection{Vision experts}
\label{sec:single-vision-encoder-comparison}
To better establish the foundation for multi-vision expert fusion, we extend the toolbox with vision experts pre-trained on different tasks and resolutions, and verify our findings on high-resolution adaptation with these experts. This also helps us identify the distinct advantages of different experts. We collect a set of vision experts, including: \textit{(1) Vision-Language Alignment:} \textit{CLIP}~\citep{radford2021learning} and \textit{ConvNeXt}~\citep{liu2022convnet} from \textit{OpenCLIP}~\citep{ilharco_gabriel_2021_5143773,schuhmann2022laionb}. \textit{(2) Object-Centric:} \textit{EVA-02}~\citep{fang2023eva,fang2023eva02} pre-trained on detection datasets. \textit{(3) OCR:} \textit{Pix2Struct}~\citep{lee2023pix2struct}.\textit{(4) Segmentation:} \textit{SAM}~\citep{kirillov2023segment}. \textit{(5) Self-supervised:} \textit{DINOv2}~\citep{oquab2023dinov2}. We resize the output 2D feature maps of each vision encoder using bilinear interpolation or pixel shuffle~\citep{pixelshffule} to ensure that the visual token number equals $1024$.

\begin{wrapfigure}{rt}{0.63\textwidth}
    \makeatletter\def\@captype{table}\makeatother
    \caption{\textbf{Comparison between different vision experts as the MLLM encoders.}}
    \label{tab:different-vision-experts}
    \vspace{-5pt}
    \small
    \centering
    \renewcommand{\arraystretch}{1.2} 
    \addtolength{\tabcolsep}{-4pt}
    \resizebox{\linewidth}{!}{
    \begin{tabular}{cccccccccccccccccc}
    \toprule[1pt]
    Category & Vision Encoder & ~~Res.~~ & Post-process & Unfreeze & ~~Avg.~~ & Model Link\\
    \midrule
    \multirow{2}{*}{\textit{VL Alignment}} & \multirow{2}{*}{\textit{ConvNeXt}} & \multirow{2}{*}{1024} & \multirow{2}{*}{None} & \ding{55} & 654.6  & \multirow{2}{*}{\href{https://huggingface.co/laion/CLIP-convnext_xxlarge-laion2B-s34B-b82K-augreg-soup}{ConvNeXt-XXL}}\\
    & & & & \ding{51} & \textbf{682.1} & \\
    \midrule
    \multirow{2}{*}{\textit{Segmentation}} & \multirow{2}{*}{\textit{SAM}} & \multirow{2}{*}{1024} & \multirow{2}{*}{Pixel Unshuffle} & \ding{55} & 486.2 & \multirow{2}{*}{\href{https://huggingface.co/facebook/sam-vit-large}{SAM-ViT-Large}} \\
    & & & & \ding{51} & \textbf{510.5} \\
    \midrule
    \multirow{2}{*}{\textit{Object Detection}} & \multirow{2}{*}{\textit{EVA-02}} & \multirow{2}{*}{1024} & \multirow{2}{*}{Resize} & \ding{55} & 543.7 & \multirow{2}{*}{\href{https://huggingface.co/Yuxin-CV/EVA-02/blob/main/eva02/det/eva02_L_coco_det_sys_o365.pth}{EVA-02-L-Det}} \\
    & & & & \ding{51} & \textbf{639.1} & \\
    \midrule
    \multirow{2}{*}{\textit{Text Recognition}} & \multirow{2}{*}{\textit{Pix2Struct}} & \multirow{2}{*}{1024} & \multirow{2}{*}{Resize} & \ding{55} & 598.6 & \multirow{2}{*}{\href{https://huggingface.co/google/pix2struct-large}{Pix2Struct-02-Large}} \\
    & & & & \ding{51} & \textbf{606.2} \\
    \midrule
    \multirow{2}{*}{\textit{Self-Supervised}} & \multirow{2}{*}{\textit{DINOv2}} & \multirow{2}{*}{448} & \multirow{2}{*}{None} & \ding{55} & 520.7 & \multirow{2}{*}{\href{https://github.com/facebookresearch/dinov2/blob/main/MODEL_CARD.md}{ViT-L/14-Reg}} \\
    & & & & \ding{51} & \textbf{537.3} & \\
    \bottomrule[1pt]
    \end{tabular}
    }
\end{wrapfigure}

Results in Table~\ref{tab:different-vision-experts} show that unfreezing the vision experts again leads to consistent improvement, which is aligned with Sec.~\ref{sec:hr-adaption}. In addition, results in Table~\ref{tab:different-vision-experts-complete} (see Appendix~\ref{sec:appendix_bench_detail}) further demonstrate that \textit{\textbf{MLLMs with these task-specific vision encoders achieve optimal performance in their pre-training domains}}. \textit{EVA-02} excels in the object hallucination evaluation benchmark POPE and general visual question answering benchmark GQA. \textit{CLIP} and \textit{ConvNeXt} perform well across all benchmarks, benefiting from their training on large-scale image-text pairs using contrastive loss. Conversely, while \textit{Pix2Struct} excels in text recognition, it shows limited capability in object recognition and general VQA tasks, like POPE and GQA. \textit{DINOv2} and \textit{SAM}, pre-trained with self-supervised learning and segmentation, struggle with text recognition tasks.

\subsection{Fusion strategy}

Existing MLLM frameworks have proposed various mixture-of-vision-encoder strategies, with the hope that their domain-specific strengths can be leveraged. In all cases, improvements in MLLM performance have been reported with the fusion of vision encoders. However, the roles of the fusion strategies as part of their MLLM architecture innovations, have not been decoupled and clearly studied under an ``apples to apples'' comparison. It is thus not entirely clear how much improvement is from the fusion strategies themselves versus the improved representations from various encoders.

We notice that existing popular fusion strategies, despite their variations in designs, can be broadly represented by the following several categories: (1) \textit{Sequence Append}: directly appending the visual tokens from different backbones as a longer sequence~\citep{fan2024mousi,kar2024brave}; (2) \textit{Channel Concatenation}: concatenating the visual tokens along the channel dimension without increasing the sequence length~\citep{lin2023sphinx,karamcheti2024prismatic}; 
(3) \textit{LLaVA-HR}: injecting high-resolution features into low-resolution vision encoders using mixture-of-resolution adapter~\citep{luo2024feast}; 
(4) \textit{Mini-Gemini}: using the \textit{CLIP} tokens as the low-resolution queries to cross-attend another high-resolution vision encoder in the co-located local windows~\citep{li2024mini}. (5) \textit{Deformable Attention}: a new baseline we introduce on top of \textit{Mini-Gemini}, where the vanilla window attention is replaced with deformable attention~\citep{deformable-detr}. Fig.~\ref{fig:workflow_figure} gives a detailed illustration of these fusion strategies. To better study them, we choose ``\textit{CLIP}+\textit{ConvNeXt}'' and ``\textit{CLIP}+\textit{ConvNeXt}+\textit{SAM}'' as the base multi-encoder combinations to perform comparisons.

\begin{wrapfigure}{rt}{0.47\textwidth}
    \vspace{-12pt}
    \makeatletter\def\@captype{table}\makeatother
    \caption{\textbf{Comparison of different fusion methods for different vision experts.} ``\#Token(V)'' denotes the number of visual tokens. ``\#Tokens/s'' denotes the inference throughtput of the whole pipeline.
    }
    \label{tab:fusion}
    \small
    \centering
    \renewcommand{\arraystretch}{1.2} 
    \addtolength{\tabcolsep}{-4pt}
    \resizebox{\linewidth}{!}{
        \begin{tabular}{c|ccccccccccccccccc}
        \toprule[1pt]
        Vision Encoders & Fusion & \#Token(V) & \#Tokens/s & \#Params & Avg. \\
        \midrule
        \multirow{5}{*}{\textit{CLIP} + \textit{ConvNeXt}} & \textit{Seq. Append} & 2048 & 46.1 & 1200M &\textbf{690.5} \\
        & \textit{Channel Concat.} & 1024 & \textbf{47.3} & \textbf{1184M} & 681.5 \\
        & \textit{LLaVA-HR} & 1024 & 47.0 & 1219M & 678.7 \\
        & \textit{Mini-Gemini} & 1024 & 45.3 & 1201M & 672.5 \\
        & \textit{Deformable Attn.} & 1024 & \textbf{47.3} & 1201M &674.3 \\
        \midrule
        \multirow{2}{*}{\begin{tabular}[c]{@{}c@{}}\textit{CLIP} + \textit{ConvNeXt} \\ + \textit{SAM}\end{tabular}} & \textit{Seq. Append} & 3072 & 40.3 & 1529M & 686.2 \\
        & \textit{Channel Concat.} & 1024 & \textbf{46.3} & \textbf{1495M} & \textbf{690.4} \\
        \bottomrule[1pt]
    \end{tabular}
    }
    \vspace{-10pt}
\end{wrapfigure}

Our study in Table~\ref{tab:fusion} shows that \textit{Channel Concatenation} stands out with the best performance, expandability, and efficiency. The ``injection-based'' methods, such as \textit{LLaVA-HR}, \textit{Mini-Gemini} and \textit{Deformable Attention}, are in general less competitive on TextVQA~\citep{singh2019towards} and OCRBench~\citep{liu2023hidden}, performing worse than using \textit{ConvNeXt} alone as the vision encoder. Although sequence append shows comparable performance to channel concatenation, it faces the challenge of handling significantly increased sequence lengths with additional vision encoders.

\subsection{Vison-language Pre-Alignment}
\label{sec:pre-alignment}

\begin{figure*}[t]
    \centering
    \includegraphics[width=1.0\linewidth]{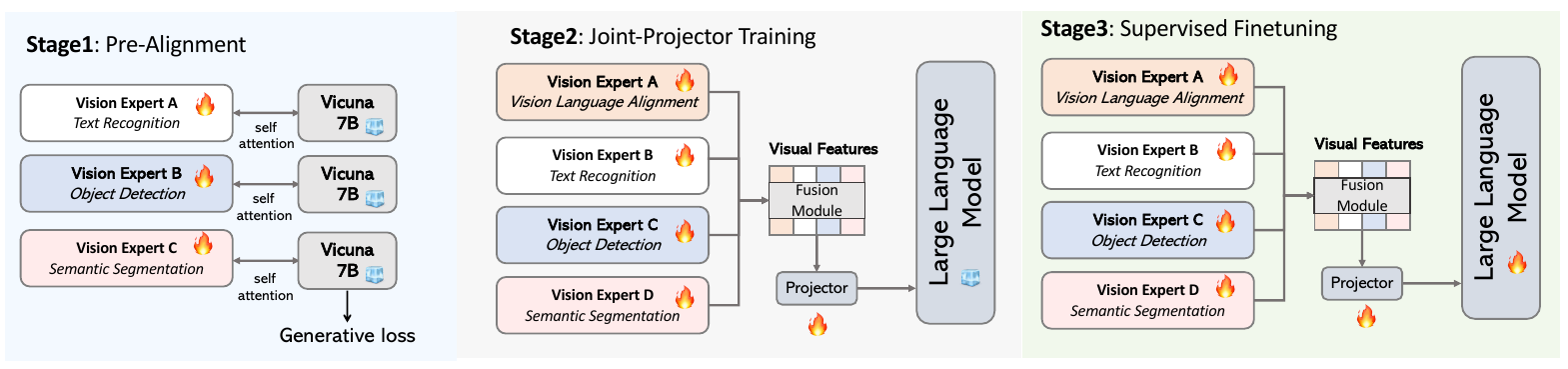}
    \caption{\textbf{The proposed training strategy of \textit{Eagle}.} It consists of three progressive stages, including \textit{vision-language pre-alignment training}, \textit{joint-project training} and \textit{supervised fine-tuning}. These stages effectively leverage public data from diverse sources, ranging from noisy image-text pairs on the web to high-quality caption, VQA, and multimodal dialogue datasets.}
    \label{fig:pre-align}
    \vspace{-0.2in}
\end{figure*}

As shown in Table~\ref{tab:different-vision-experts}, encoders pre-trained exclusively on vision tasks (\textit{e.g.}, detection, OCR, and segmentation) are less competitive compared to those pre-trained on vision language alignment. This is possibly due to representational inconsistencies when integrated with large language models. Additionally, when combining different encoders, there is a gap between these encoders, creating difficulties in the training process. To address this feature inconsistency, we propose a \textit{Pre-Alignment} training stage that first aligns each individual vision encoder with the same large language model, fostering better synergy between visual and linguistic capabilities.

Fig.~\ref{fig:pre-align} depicts our pre-alignment strategy. Instead of training a projector to simultaneously align multiple vision experts as in \textit{LLaVA}'s~\citep{liu2023improved} original pre-training strategy, we first align the representation of each individual expert with a smaller language model (Vicuna-7B in practice) using next-token-prediction supervision. As shown in Fig.~\ref{fig:pre-align}, with pre-alignment, the whole training process consists of three steps: 1) \textit{training each pre-trained vision expert with their own projector,  while keeping the language model frozen;} 2) \textit{combining all vision experts from the first step and training both the projector and vision experts;} 3) \textit{training the whole model on SFT data.}

\begin{wrapfigure}{rt}{0.45\textwidth}
    \vspace{-10pt}
    \makeatletter\def\@captype{table}\makeatother
    \caption{\textbf{The effectiveness of \textit{Pre-alignment}}.
    }
    \vspace{-4pt}
    \label{tab:pre-align}
    \small
    \centering
    \renewcommand{\arraystretch}{1.2} 
    \addtolength{\tabcolsep}{-4pt}
    \resizebox{\linewidth}{!}{
        \begin{tabular}{ccccc}
        \toprule[1pt]
        CLIP & Vision Expert (X)  & Unfreeze & Pre-align  & Avg.  \\
        \midrule
        \multirow{4}{*}{\textit{CLIP-448}} & \multirow{4}{*}{\textit{SAM-1024}} & \ding{55} & \ding{55} & 630.6\\
        & & \ding{55} & \ding{51} & 648.5\\
        & & \ding{51} & \ding{55} & 662.5\\
        & & \ding{51} & \ding{51} & \textbf{672.3}\\
        \hline
        \multirow{4}{*}{\textit{CLIP-448}} & \multirow{4}{*}{\textit{ConvNext-1024}} & \ding{55} & \ding{55} & 652.0\\
        & & \ding{55} & \ding{51} & 670.1\\
        & & \ding{51} & \ding{55} & 681.5\\
        & & \ding{51} & \ding{51} & \textbf{686.2}\\
        \hline
        \multirow{4}{*}{\textit{CLIP-448}} &\multirow{4}{*}{\textit{Pix2Struct-1024}} & \ding{55} & \ding{55} & 653.5\\
        & & \ding{55} & \ding{51} & 665.7\\
        & & \ding{51} & \ding{55} & 673.7\\
        & & \ding{51} & \ding{51} & \textbf{680.4}\\
        \hline
        \multirow{4}{*}{\textit{CLIP-448}} &\multirow{4}{*}{\textit{EVA-02-L-1024}} & \ding{55} & \ding{55} & 630.2\\
        & & \ding{55} & \ding{51} & 645.2\\
        & & \ding{51} & \ding{55} & 659.2\\
        & & \ding{51} & \ding{51} & \textbf{668.2}\\
        \bottomrule[1pt]
        \end{tabular}
    }
    \vspace{-35pt}
\end{wrapfigure}

To verify the proposed method, we compare the pre-alignment strategy with the normal two-stage training strategy in Table~\ref{tab:pre-align}, considering both freezing and unfreezing vision experts for comparison. As shown in Table~\ref{tab:pre-align}, although unfreezing the vision experts during SFT helps improve performance by updating the vision experts to fit the language model, the \textit{Pre-Align} strategy more effectively mitigates the inherent biases of each vision expert and stabilizes the training process, subsequently improving overall performance.

\subsection{Extension to multi-experts}
\label{sec:combining-multiple-vision-encoders}
With the optimized strategies and training recipes of incorporating individual vision experts, we consider the incorporation of even more vision experts to push the limit. To conduct the search in a systematic manner, we adopt a step-by-step greedy strategy to incorporate additional vision experts.

\begin{figure}[!t]
    \centering
    \begin{minipage}[!t]{0.51\textwidth}
        \centering
        \small
        \renewcommand{\arraystretch}{1.2} 
        \addtolength{\tabcolsep}{-3pt}
        \resizebox{1.1\textwidth}{!}{
        \begin{tabular}{c|lccccc}
        \toprule[1pt]
        \#Encoder          & Encoder Combination & Config & \#Params (M) & FLOPs (G) & Img/Sec & Avg. \\
                           \midrule
        2                  & \textit{CL} + \textit{CN} & X2 & 1155.2 & 3347.2 & 18.1 & 681.5 \\
                           \midrule
        \multirow{4}{*}{3} & \textit{CL} + \textit{CN} + \textit{DI} &    & 1460.6 & 3659.9 & 15.1 & 685.4 \\
                           & \textit{CL} + \textit{CN} + \textit{SA} &    & 1463.9 & 4657.8 & 8.8 & 690.4 \\
                           & \textit{CL} + \textit{CN} + \textit{PS} &    & 1669.6 & 4373.2  & 6.9 & 685.1 \\
                           & \textit{CL} + \textit{CN} + \textit{EV} & X3 & 1459.6 & 4280.9 & 9.1 & \textbf{690.7} \\
                           \midrule
        \multirow{3}{*}{4} & \textit{CL} + \textit{CN} + \textit{EV} + \textit{DI} &    & 1765.1 & 4593.6 & 8.3 & 688.0 \\
                           & \textit{CL} + \textit{CN} + \textit{EV} + \textit{SA} &    & 1768.4 & 5591.5 & 5.9 & 689.4 \\
                           & \textit{CL} + \textit{CN} + \textit{EV} + \textit{PS} & X4 & 1974.1 & 5306.9 & 5.0 & \textbf{694.6} \\
                           \midrule
        \multirow{2}{*}{5} & \textit{CL} + \textit{CN} + \textit{EV} + \textit{PS} + \textit{DI} &    & 2279.5 & 5619.5 & 4.7 & 684.7 \\
                           & \textit{CL} + \textit{CN} + \textit{EV} + \textit{PS} + \textit{SA} & X5 & 2282.8 & 6617.4 & 3.8 & \textbf{697.1} \\
                           \midrule
        6                  & \textit{CL} + \textit{CN} + \textit{EV} + \textit{PS} + \textit{SA} + \textit{DI} & X6 & 2588.2 & 6930.1 & 3.6 & 686.8 \\
        \bottomrule[1pt]
        \end{tabular}
        }
    \end{minipage}
    \hfill
    \begin{minipage}[!t]{0.42\textwidth}
        \centering
        \includegraphics[width=\textwidth]{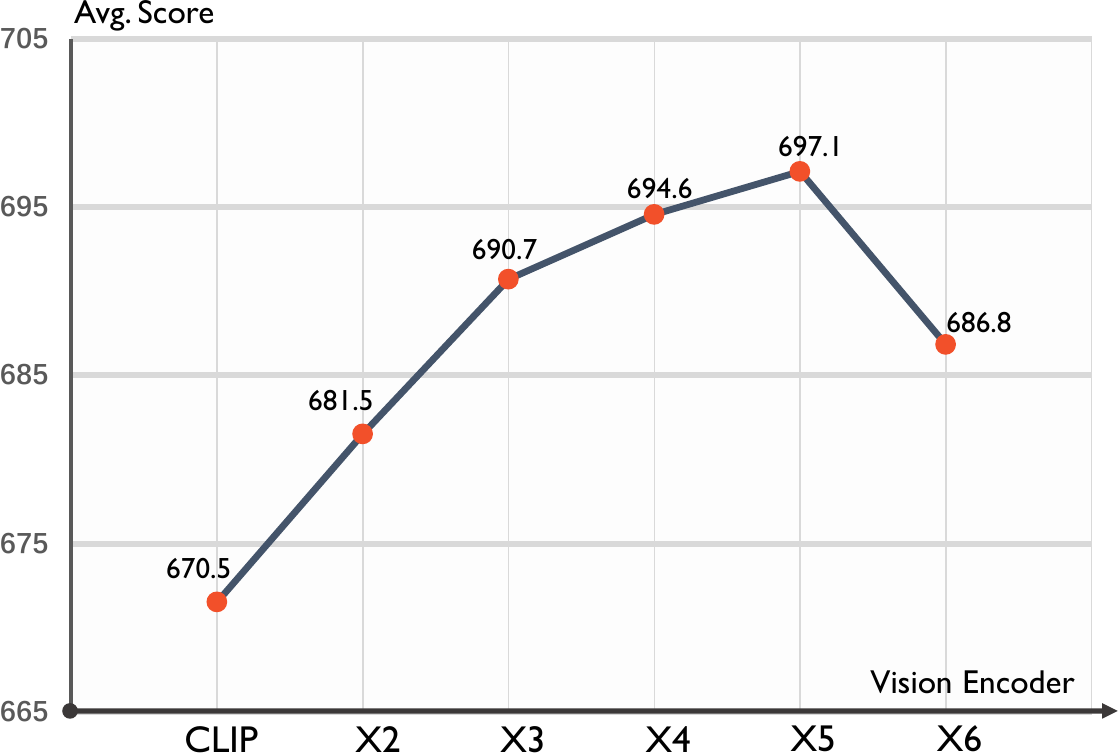}
        \label{fig:vis-encoder-ablation}
        \vspace{-10pt}
    \end{minipage}
    \caption{\textbf{Results of vision expert selection process.} \textit{CL}, \textit{CN}, \textit{EV}, \textit{PS}, \textit{SA} and \textit{DI} denote \textit{CLIP}, \textit{ConvNeXt}, \textit{EVA-02}, \textit{Pix2Struct}, \textit{SAM} and \textit{DINOv2}, respectively. (Left) The performance of various vision encoder combinations is presented, highlighting how different pairings influence overall effectiveness. ``\#Parames'', ``FLOPs'' and ``Img/Sec'' denote the model size, complexity and throughput (bs=4) of the vision encoder. (Right) The curve illustrates the average score as the number of vision encoders increases. Each point on the curve represents the best-performing combination for the corresponding number of vision encoders. No \textit{Pre-Alignment} is used in this comparison.}
    \label{tab:multiple-vision-encoders-results}
\end{figure}

We consider the vision experts discussed in Section~\ref{sec:single-vision-encoder-comparison} for experiments. We mark \textit{CLIP}, \textit{ConvNeXt}, \textit{SAM}, \textit{DINOv2}, \textit{Pix2Struct}, and \textit{EVA-02-L} as \textit{CL}, \textit{CN}, \textit{SA}, \textit{DI}, \textit{PS}, and \textit{EV}, respectively. A round-robin scheme, as shown in Fig.~\ref{tab:multiple-vision-encoders-results}, is adopted. We first use the two top-performing vision encoders, \textit{CLIP} and \textit{ConvNeXt}, as the basis and gradually add one more vision encoder each time. In each round, the best-performing vision encoder combination is retained for the next round.

Fig.~\ref{tab:multiple-vision-encoders-results} reveals several insights. Generally, \textbf{introducing additional vision encoders enhances the performance}. This indicates that the distinct advantages of different encoders can be preserved and utilized; for example, integrating the \textit{EVA-02} encoder improves metrics on the POPE benchmark. Although individual metrics may vary, the aggregated performance shows an upward trend, as evidenced by normalized average metrics, suggesting that the overall efficacy of the system is enhanced with more encoders. Also, Fig.~\ref{tab:multiple-vision-encoders-results} shows that the best combination of vision experts are \textit{CLIP}, \textit{ConvNeXt}, \textit{SAM}, \textit{Pix2Struct}, and \textit{EVA-02}. We will use this recipe in the final model.

\section{Experiments}
\label{sec:experiment}
In this section, we take the findings and the best-explored design from Section~\ref{sec:preliminary} and compare them against the current state-of-the-art MLLMs on different tasks.

\subsection{Implementation details}
\noindent \textbf{Language models.} We use Vicuna-v1.5-7B~\citep{vicuna2023}, Llama3-8B~\citep{llama3modelcard} and Vicuna-v1.5-13B~\citep{vicuna2023} as the language models. 

\noindent \textbf{Vision encoders.} We follow the best X4 and X5 configurations, where the interpolated \textit{CLIP-448} and pre-aligned vision experts are channel-concatenated, and trained following the exact best training recipes in Figure~\ref{fig:pre-align} and Table~\ref{tab:pre-align}.

\noindent \textbf{Training recipe.} On Eagle1.8M, we follow the base recipe in Section~\ref{sec:baseline-setup} with encoder learning rate the same as SFT (2e-5). On Cambrian data, we follow~\citet{tong2024cambrian} with PT/SFT bs=1024.

\subsection{Main results}

\textbf{Evaluation on visual question answering tasks.}
We compare \textit{Eagle} model series across three Visual Question Answering (VQA) benchmarks, including GQA~\citep{hudson2019gqa}, VQAv2~\citep{balanced_vqa_v2} and VizWiz~\citep{2018vizwiz}. As shown in Table~\ref{tab:mllm-results}, \textit{Eagle-X5} achieves state-of-the-art performance on GQA and VQAv2, underscoring the advantages with additional vision experts.

\begin{table}[!t]
    \vspace{-0.1in}
    \caption{\textbf{Main results with base training data.} SQA$^\mathrm{I}$ denotes image split of ScienceQA.}
    \label{tab:mllm-results}
    \small
    \centering
    \renewcommand{\arraystretch}{1.2} 
    \addtolength{\tabcolsep}{-1.5pt}
    \resizebox{\textwidth}{!}{
    \begin{tabular}{c|l|cccccccccccccc}
    \toprule[1.5pt]
        & \multicolumn{1}{c|}{Model} & \rotatebox{90}{MME} & \rotatebox{90}{MMB} & \rotatebox{90}{SEED} & \rotatebox{90}{MathVista} & \rotatebox{90}{MMMU} & \rotatebox{90}{POPE} & \rotatebox{90}{SQA$^\mathrm{I}$} & \rotatebox{90}{GQA} & \rotatebox{90}{VizWiz} & \rotatebox{90}{VQAv2} & \rotatebox{90}{OCR} & \rotatebox{90}{TextVQA} & \rotatebox{90}{ChartQA} \\
        \midrule
        \multicolumn{1}{c|}{\multirow{8}{*}{\rotatebox{90}{\textit{Vicuna-7B} \& \textit{Qwen-7B}}}} &
        \textit{LLaVA-1.5}~\citep{liu2023improved} & 1510 & 64.3 & 58.6 & - & - & 85.9 & 66.8 & 62.0* & 50.0 & 78.5* & 297 & 58.2 & - \\
        & \textit{LLaVA-NeXt}~\citep{liu2024llavanext} & 1519 & 67.4 & 70.2 & 34.6 & 35.8 & 86.5 & 70.1 & 64.2* & 57.6 & 80.0* & 490 & 64.9 & - \\
        & \textit{InternVL}~\citep{chen2023internvl} & 1525 & - & 65.4 & - & - & 86.4 & - & 62.9* & 52.5 & 79.3* & - & 57.0 & - \\
        & \textit{LLaVA-HR}~\citep{luo2024feast} & 1554 & - & 64.2 & - & - & 87.6 & 65.1 & 64.2* & 48.7 & 81.9* & - & 67.1 & - \\
        & \textit{Monkey}~\citep{li2024monkey} & - & - & - & - & - & - & - & 60.7* & \textbf{61.2}* & 80.3* & 514 & 67.6 & 65.1 \\
        & \textit{Mini-Gemini}~\citep{li2024mini} & 1523 & 65.8 & - & 32.2 & 36.8 & - & 71.1 & 64.5* & - & - & 477 & 65.2 & - \\
        \cline{2-15}
        & \textit{Eagle-X5} & 1528 & 68.4 & \textbf{73.9} & 37.0 & 36.3 & \textbf{88.8} & 70.0 & \textbf{64.9}* & 54.4 & 83.4* & 529 & 71.2 & 67.7 \\ 
        & \textit{Eagle-X5 (+Pre-Align)} & \textbf{1582} & \textbf{69.7} & 73.7 & \textbf{38.2} & \textbf{38.0} & 88.7 & \textbf{71.9} & 64.6* & 58.7 & \textbf{83.6*}& \textbf{566} & \textbf{71.9} & \textbf{69.3} \\
        \midrule 
        \multicolumn{1}{c|}{\multirow{10}{*}{\rotatebox{90}{\textit{Vicuna-13B}}}} & \textit{LLaVA-1.5}~\citep{liu2023improved} & 1531 & 67.7 & 61.6 & - & 36.4 & 85.9 & 71.6 & 63.3* & 53.6 & 80.0* & 331 & 61.3 & - \\
        & \textit{LLaVA-NeXt}~\citep{liu2024llavanext} & 1575 & 70.0 & 71.9 & 35.3 & 36.2 & 86.2 & \textbf{73.5} &  65.4* & 60.5 & 82.8* & 514 & 67.1 & 62.2 \\
        & \textit{InternVL}~\citep{chen2023internvl} & 1546 & - & - & - & - & 87.1 & -  &  63.9* & 54.6 & 80.2* & 517 & 58.7 & - \\
        & \textit{LLaVA-UHD}~\citep{xu2024llava-uhd} & 1535 & 68.0 & - & - & - & 89.1 & 72.0 & 65.2* & 56.1 & 81.7* & - & 67.7 & - \\
        & \textit{LLaVA-HR}~\citep{luo2024feast} & 1540 & - & 64.5 & - & - & 87.8 & 68.1 & 64.8* & 57.9 & 82.6* & - & 68.1 & - \\
        & \textit{Mini-Gemini}~\citep{li2024mini} & 1565 & 68.6 & 70.6 & 37.0 & 37.3 & - & 71.9 & 65.8* & - & - & 466 & 65.9 & 56.6 \\
        \cline{2-15}
        & \textit{Eagle-X5} & \textbf{1609} & 69.2 & 74.1 & 38.8 & 36.6 & 87.8 & 72.7 & \textbf{66.2*} & 59.3 & 83.8* & 574 & \textbf{74.2} & 69.9 \\ 
        & \textit{Eagle-X5 (+Pre-Align)} & 1605 & \textbf{71.6} & \textbf{74.9} & \textbf{42.7} & \textbf{38.5} & \textbf{89.2} & \textbf{75.5} & 64.6* & \textbf{60.9} & \textbf{84.5}* & \textbf{598} & 73.3 & \textbf{72.1} \\ 
        \bottomrule[1.5pt]
    \end{tabular}
    }
    \vspace{-0.1in}
\end{table}

\textbf{Evaluation on OCR and chart understanding tasks.}
To evaluate the OCR, document, and chart understanding capabilities of \textit{Eagle}, we benchmark our model on OCRBench~\citep{liu2023hidden}, TextVQA~\citep{singh2019towards}, and ChartQA~\citep{masry2022chartqa}. As illustrated in Table~\ref{tab:mllm-results}, our model significantly surpasses competitors on TextVQA, benefiting from its high-resolution architecture and integration of different vision encoders. Notably, \textit{Eagle} maintains a straightforward design, supporting up to 1024x1024 resolution without requiring complex tile decomposition of images.

\begin{figure*}[t]
    \centering
      \includegraphics[width=0.97\textwidth]{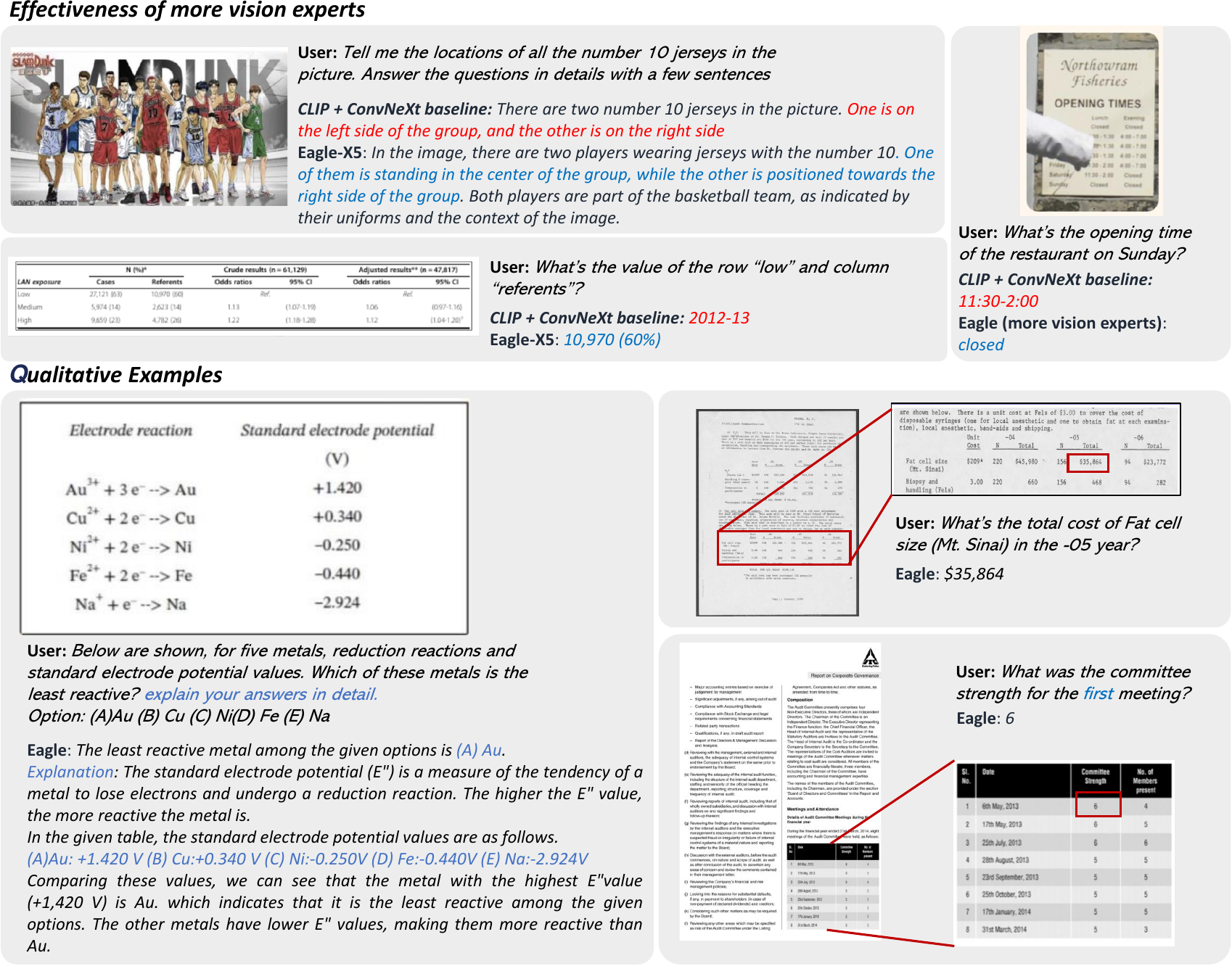}
    \caption{
    \textbf{Qualitative comparison of different numbers of vision experts.} \textbf{\textit{Baseline}} means \textit{Eagle} with only \textit{CLIP}+\textit{ConvNext}. \textbf{\textit{More Vision Experts}} denotes the \textit{Eagle-X5} model. We highlight a failure case in \textcolor{red}{RED}. \textcolor{blue}{BLUE} indicates the correct answers. With more vision experts, \textit{Eagle} can more precisely capture the information in the image and provide the correct answer.
    }\label{fig:qualitative-examples_encoders}
    \vspace{-0.1in}
\end{figure*}

Fig.~\ref{fig:qualitative-examples_encoders} shows some examples of OCR and document understanding cases. With high-resolution adaptation and more vision experts, our model can identify small text within images and accurately extract information according to the users' instructions. To better understand the benefits of introducing experts pre-trained on other vision tasks, we visualize the results of a model with only the \textit{ConvNeXt} and \textit{CLIP} vision encoders, compared to the results of \textit{Eagle-X5} in Fig.~\ref{fig:qualitative-examples_encoders}. With the full set of vision encoders, the model can successfully correct mistakes, showing that even when equipped with high-resolution vision encoders pre-trained on vision-language alignment, the model's abilities can still be enhanced by integrating additional vision experts pre-trained on diverse vision tasks.

\textbf{Evaluation on multimodal benchmarks.}
We evaluate \textit{Eagle} on seven benchmarks for MLLMs to demonstrate its capabilities from different perspectives, including MME~\citep{fu2023mme}, MMBench~\citep{MMBench}, SEED~\citep{li2023seed}, MathVista~\citep{lu2024mathvista}, MMMU~\citep{yue2024mmmu}, ScienceQA~\citep{saikh2022scienceqa}, and POPE~\citep{li2023evaluating}. Specifically, MME, MMBench, and SEED assess the overall performance on various real-world tasks based on reasoning, recognition, knowledge, and OCR. MMMU focuses on challenging problems from diverse domains that require college-level knowledge. POPE evaluates the visual hallucinations of MLLMs. The metrics used in our paper adhere to the default settings of these benchmarks.
We report the perception score for MME, the \texttt{en\_dev} split for MMBench, the \texttt{image} split of SEED, the \texttt{test-mini} split of MathVista, the \texttt{val} split of MMMU, the F1-score of POPE, and the \texttt{image} split of SQA to align with the reported scores from other models.

\begin{wrapfigure}{rt}{0.65\textwidth}
    \vspace{-12pt}
    \makeatletter\def\@captype{table}\makeatother
    \caption{\textbf{Comparison between different training strategies.} ``1 epoch'' means we train \textit{Eagle} for 1 epoch in the supervised fine-tuning stage. `unlock*'' means we unlock vision encoders in the pre-training stage. More details in Table. \ref{tab:training_stratgy_more}.
    }
    \vspace{-4pt}
    \label{tab:pre-align-step}
    \small
    \centering
    \renewcommand{\arraystretch}{1.5} 
    \addtolength{\tabcolsep}{-2pt}
    \resizebox{1.0\linewidth}{!}{
        \begin{tabular}{lcccc}
        \toprule[1.5pt]
        Config Summary  & Pre-align & Pre-train  & Fine-tune& Avg.  \\
        \midrule
        1 epoch & \ding{55} & LLaVA-595K & Eagle1.8M & 697.1\\
        2 epoch & \ding{55} & LLaVA-595K & Eagle1.8M & 698.3\\
        1 epoch, unlock* & \ding{55} & LLaVA-595K & Eagle1.8M & 698.0\\
        1 epoch, unlock* & \ding{55} & LLaVA-595K+Eagle1.8M & Eagle1.8M & \textbf{699.5}\\
        \midrule
        1 epoch & Eagle1.8M & LLaVA-595K & Eagle1.8M & 706.6\\
        1 epoch, unlock* & Eagle1.8M & LLaVA-595K & Eagle1.8M & 707.1\\
        1 epoch, unlock* & LLaVA-595K+Eagle1.8M & LLaVA-595K & Eagle1.8M & 707.8\\
        1 epoch, unlock* & LLaVA-595K+Eagle1.8M & LLaVA-595K+Eagle1.8M & Eagle1.8M & \textbf{708.9}\\
        \bottomrule[1.5pt]
        \end{tabular}}
    \vspace{-10pt}
\end{wrapfigure}

From the data presented in Table~\ref{tab:mllm-results}, \textit{Eagle} consistently surpasses existing models across various MLLMs on SEED and MME, demonstrating the comprehensive knowledge and reasoning abilities of \textit{Eagle}. With the help of vision encoders on object-centric tasks, \textit{Eagle} also achieves the best performance on the POPE benchmark. Additionally, the \textit{Pre-Alignment} strategy discussed in Sec.~\ref{sec:pre-alignment} has been found to further enhance performance when integrating multiple task-specific vision backbones. This approach not only mitigates the inherent biases of each vision expert and the synergy between different modalities but also establishes a robust framework for multiple-expert fusion.

\textbf{Study on more advanced training recipes.} Table.~\ref{tab:pre-align-step} presents our step-by-step experiments to study the training recipes. We found that the best recipe is to first pre-align each vision expert on \textit{LLaVA-595K} + \textit{Eagle1.8M}. In the pretraining stage, we combine all vision experts from the first step and training both the projector and vision experts on \textit{LLaVA-595K} + \textit{Eagle1.8M}. Finally, we train the whole model on the \textit{Eagle1.8M}.

\textbf{Comparison with \textit{Cambrian-1}.}
Using the same pre-training and supervised fine-tuning datasets from \textit{Cambrian-1}~\citep{tong2024cambrian}, \textit{Eagle} demonstrates superior performance across all the evaluated benchmarks without bells and whistles. As shown in Table~\ref{tab:vqa-results-cambrain}, \textit{Eagle} outperforms the \textit{Cambrian-1} counterparts considerably for the \textit{OCR and Chart} category. Consistent improvements are also observed for the \textit{General}, \textit{Knowledge}, and \textit{Vision-Centric} categories, showing the robustness and generalization ability of the improved perception design in \textit{Eagle}.

\begin{table}[!t]
\caption{\textbf{Results using the same training data as \textit{Cambrian-1}}~\citep{tong2024cambrian}. SQA$^\mathrm{I}$ denotes ScienceQA-IMG~\citep{saikh2022scienceqa}. RWQA denotes the RealworldQA~\citep{RWQA}.
}
\label{tab:vqa-results-cambrain}
\small
\centering
\renewcommand{\arraystretch}{1.2} 
\addtolength{\tabcolsep}{-2pt}
\resizebox{\textwidth}{!}{
\begin{tabular}{r|ccccc|ccccc|ccccc|ccc}
\toprule[1.5pt]
\multicolumn{1}{c}{Model} & \multicolumn{5}{c}{Knowledge} & \multicolumn{5}{c}{General} & \multicolumn{5}{c}{OCR and Chart} & \multicolumn{3}{c}{Vision-Centric} \\
\midrule
&  \rotatebox{90}{Avg} & \rotatebox{90}{SQA$^{\mathrm{I}}$} & \rotatebox{90}{MMMU} & \rotatebox{90}{MathVista} & \rotatebox{90}{AI2D} & \rotatebox{90}{Avg} & \rotatebox{90}{MME} & \rotatebox{90}{MMB} & \rotatebox{90}{SEED} & \rotatebox{90}{GQA} & \rotatebox{90}{Avg} & \rotatebox{90}{ChartQA} & \rotatebox{90}{OCR} & \rotatebox{90}{TextVQA} & \rotatebox{90}{DocVQA} & \rotatebox{90}{Avg} & \rotatebox{90}{MMVP} & \rotatebox{90}{RWQA} \\
\hline
 \rowcolor{mygray}\textit{Llama3-8B} & & & & & & & & & & & & & & & & & & \\
\textit{MGM-HD} & 55.7 & 75.1 & 37.3 & 37.0 & 73.5 & 72.7 & \textbf{1606} & 72.7 & 73.2 & 64.5 & 62.9 & 59.1 & 47.7 & 70.2 & 74.6 & 40.4 & 18.7 & 62.1  \\
\textit{Cambrian-1} & 61.3 & 80.4 & 42.7 & 49.0 & 73.0 & 73.1 & 1547 & \textbf{75.9} & 74.7 & 64.6 & 71.3 & 73.3 & 62.4 & 71.7 & 77.8 & 57.6 & 51.3 & 64.2  \\
\textit{Eagle-X5} & \textbf{65.2} & \textbf{84.1} & \textbf{43.5} & \textbf{56.9} & \textbf{76.2} & \textbf{74.0} & 1587 & 75.5 & \textbf{76.5}& \textbf{64.9} & \textbf{77.0} & \textbf{80.7} & \textbf{62.6} & \textbf{76.7} & \textbf{87.1} & \textbf{59.6}& \textbf{52.0} & \textbf{67.2} \\
\hline
 \rowcolor{mygray}\textit{Vicuna-13B} & & & & & & & & & & & & & & & & & & \\ 
\textit{MGM-HD} & 54.1 & 71.9 & 37.3 & 37.0 & 70.1 & 70.7 & 1597 & 68.6 & 70.6 & 63.7 & 60.8 & 56.6 & 46.6 & 70.2 & 69.8 & 38.4 & 19.3 & 57.5 \\
\textit{Cambrian-1} & 60.2 & 79.3 & 40.0 & 48.0 & 73.6 & 73.7 & 1610 & \textbf{75.7} & 74.4 & 64.3 & 71.3 & 73.8 & 61.9 & 72.8 & 76.8 & 52.2 & 41.3 & 63.0 \\
\textit{Eagle-X5} & \textbf{63.8} & \textbf{82.6} & \textbf{42.2} & \textbf{54.6} & \textbf{73.8} & \textbf{74.6} & \textbf{1651} & \textbf{75.7} & \textbf{75.0} & \textbf{65.0} & \textbf{75.7} & \textbf{78.6} & \textbf{62.4} & \textbf{74.9} & \textbf{86.7} & \textbf{54.8}& \textbf{44.6} & \textbf{65.0} \\
\hline
 \rowcolor{mygray}\textit{Yi-34B} & & & & & & & & & & & & & & & & & & \\
\textit{MGM-HD} & 62.4 & 77.7 & 48.0 & 43.4 & 80.5 & 76.2 & 1659 & 80.6 & 75.3 & 65.8 & 68.1 & 67.6 & 51.8 & 74.1 & 78.9 & 52.3 & 37.3 & 67.2 \\
\textit{Cambrian-1} & 67.0 & \textbf{85.6} & 49.7 & 53.2 & \textbf{79.7} & \textbf{76.8} & \textbf{1689} & \textbf{81.4} & 75.3 & \textbf{65.8} & 71.9 & 75.6 & 60.0 & 76.7 & 75.5 & \textbf{60.3} & \textbf{52.7} & 67.8 \\
\textit{Eagle-X5} & \textbf{68.6} & 85.5 & \textbf{53.2} & \textbf{57.9} & 79.1 & 76.3 & 1677 & 81.0 & \textbf{75.6} & 64.9 & \textbf{75.4} & \textbf{77.2} & \textbf{62.4} & \textbf{78.8} & \textbf{83.0} & 59.8& 50.0 & \textbf{69.5} \\
\bottomrule[1.5pt]
\end{tabular}
}
\vspace{-0.1in}
\end{table}

\section{Related work}

\textbf{Multimodal large language models.}
Our work is related to the general architecture design of multimodal large language models. Besides the line of representative open-source research mentioned in the introduction section, other notable families of MLLMs include, but are not limited to \textit{MiniGPT-4}~\citep{zhu2023minigpt,chen2023minigpt}, \textit{Lynx}~\citep{zeng2023matters}, \textit{Otter}~\citep{li2023otter,li2023mimic}, \textit{Qwen-VL}~\citep{bai2023qwen}, \textit{CogVLM}~\citep{wang2023cogvlm,hong2024cogagent}, \textit{VILA}~\citep{lin2023vila}, \textit{GPT-4V}~\citep{achiam2023gpt}, \textit{Gemini}~\citep{team2023gemini}, and \textit{Llama 3.1}~\citep{dubey2024llama}. Depending on how vision signals are integrated into the language model, MLLMs can be broadly categorized into ``cross-modal attention'' ones and ``prefix-tuning'' ones~\citep{yin2024survey}. The former injects visual information into different layers of LLMs using cross-modal attention~\citep{alayrac2022flamingo,li2023otter}, whereas the latter views the visual tokens as part of the language token sequence and directly append them with text embeddings~\citep{liu2023visual,li2022blip,li2023blip}. Our model belongs to the prefix-tuning family by following a \textit{LLaVA}-styled multimodal architecture. Considering that MLLM is a fast-growing field, readers are recommended to refer to more detailed studies and surveys such as~\cite{yin2024survey,li2024multimodal,fei2024multimodal}.

\textbf{Vision encoder design for MLLMs.}
Our work is closely related to research focusing on improved vision encoder designs for MLLM. Early works~\citep{liu2023visual,li2022blip} usually adopted vision encoder pre-trained on vision-language alignment tasks such as \textit{CLIP}~\citep{radford2021learning} and \textit{EVA-CLIP}~\citep{sun2023eva}. Stronger vision encoders such as \textit{SigLIP}~\citep{siglip} and \textit{InternVL}~\citep{chen2023internvl} have been proposed to improve vision-language tasks with improved designs, larger model sizes, and better training recipes. Considering that the models are often pre-trained on low-resolution images and lack the ability to encode fine-grained details, higher resolution adaptation is often performed to increase the MLLM input resolution~\citep{chen2023palix,chen2023pali3,beyer2024paligemma,chen2024far}.

Besides higher resolution adaptation, models such as \textit{LLaVA-NeXT}~\citep{liu2024llavanext}, \textit{LLaVA-UHD}~\citep{xu2024llava-uhd}, \textit{Monkey}~\citep{li2024monkey}, \textit{InternLM-XComposer}~\citep{dong2024internlm2hd}, and \textit{InternVL}~\citep{chen2024far} use adaptive tiling to divide high-resolution input into lower resolution patches that are processed separately. Although the ability to handle higher resolution is similarly made possible with the introduction of additional vision experts, its spirit is slightly orthogonal to the tiling techniques in terms of diversifying the visual representations. Both techniques are compatible and can be combined together.

Our work is most related to existing models using multiple vision encoders for improved perception. \textit{Mini-Gemini}~\citep{li2024mini} and \textit{LLaVA-HR}~\citep{luo2024feast} propose to fuse high-resolution visual features into the low-resolution visual tokens. Apart from the resolution issue, these pre-trained vision encoders may lack specific abilities such as reading text and localizing objects. Hence, a series of works have integrated vision models pre-trained on different tasks for more comprehensive capabilities. For example, \textit{Mousi}~\citep{fan2024mousi}, \textit{Prismatic VLM}~\citep{karamcheti2024prismatic} and \textit{Brave}~\citep{kar2024brave} fuse visual tokens from different vision encoders by concatenating along the channel or token direction. There are also more complex approaches, including knowledge distillation~\cite{radio}, augmenting the input prompt with the information output by the vision experts~\citep{lee2024moai,he2024incorporating,liu2024prismer}, or using a routing network to assign input to proper vision experts~\citep{zong2024mova}. In particular, \textit{Prismatic VLM}~\citep{karamcheti2024prismatic} systematically explores the design space of MLLMs across various dimensions, including data, training recipes, and notably. It also includes the vision encoder ensemble as part of its enhancement strategies. However, it lacks a comprehensive ablation study and discussion addressing the challenges associated with combining multiple vision encoders.

\section{Conclusion}
\label{sec:conclude}

We conduct an in-depth analysis study on the design space for integrating vision encoders for multimodal large language models. Unlike previous works that focus on designing novel fusion paradigms, we find systematic design choice matters and discover a series of useful techniques. Step by step, we optimize the training recipes of individual vision encoders, identify an extendable and efficient fusion method, and gradually combine vision encoders with different domain knowledge. The results show the importance of basic design space. We hope our work can serve as a new basis and bring new inspiration for the vision encoder design for MLLM.

\section{Acknowledgments}
\label{sec:acknowledgments}
The team would like to thank Yunhao Fang and Jason Lu for sharing the data and training recipes, and Yawen Luo for the assistance on figure editing. We thank Wei Ping, Zhuolin Yang, Wenliang Dai, Nayeon Lee, Boxin Wang, Ilia Karmanov, Lukas Voegtle, Philipp Fischer, Matthieu Le and Tuomas Rintamaki for their assistance on the internal codebases. We also thank the valuable discussions and input from Zhiqi Li, Guo Chen, Shilong Liu, Jihao Liu, Ming-Chang Chiu, Yunze Man, Shiyi Lan, Nadine Chang, Maying Shen, Vibashan VS, Jenny Schmalfuss, Jose Alvarez, Amala Sanjay Deshmukh, Mike Ranzinger, Greg Heinrich, Pavlo Molchanov, Vidya Murali, Parthasarathy Sriram, Mohammad Shoeybi, Song Han, Ofri Masad, Osvald Nitski, Qing Miao, Yao Xu, Jane Scowcroft, Dmitry Chichkov and Padmavathy Subramanian. Finally, the team would like to thank the Hugging Face Team that for the support of ZERO GPU demo, and the NVIDIA infrastructure team for their prompt and helpful assistance. Min Shi is partly supported by NSF Award \#2427478 and \#2229873.

\bibliography{ref}

\begin{thebibliography}{91}
\providecommand{\natexlab}[1]{#1}
\providecommand{\url}[1]{\texttt{#1}}
\expandafter\ifx\csname urlstyle\endcsname\relax
  \providecommand{\doi}[1]{doi: #1}\else
  \providecommand{\doi}{doi: \begingroup \urlstyle{rm}\Url}\fi

\bibitem[lai(2023)]{laion-gpt4v}
{LAION-GPT4v} dataset.
\newblock \url{https://huggingface.co/datasets/laion/gpt4v-dataset}, 2023.

\bibitem[Achiam et~al.(2023)Achiam, Adler, Agarwal, Ahmad, Akkaya, Aleman,
  Almeida, Altenschmidt, Altman, Anadkat, et~al.]{achiam2023gpt}
Josh Achiam, Steven Adler, Sandhini Agarwal, Lama Ahmad, Ilge Akkaya,
  Florencia~Leoni Aleman, Diogo Almeida, Janko Altenschmidt, Sam Altman,
  Shyamal Anadkat, et~al.
\newblock {GPT-4} technical report.
\newblock \emph{arXiv:2303.08774}, 2023.

\bibitem[AI@Meta(2024)]{llama3modelcard}
AI@Meta.
\newblock Llama 3 model card, 2024.
\newblock URL
  \url{https://github.com/meta-llama/llama3/blob/main/MODEL_CARD.md}.

\bibitem[Alayrac et~al.(2022)Alayrac, Donahue, Luc, Miech, Barr, Hasson, Lenc,
  Mensch, Millican, Reynolds, et~al.]{alayrac2022flamingo}
Jean-Baptiste Alayrac, Jeff Donahue, Pauline Luc, Antoine Miech, Iain Barr,
  Yana Hasson, Karel Lenc, Arthur Mensch, Katherine Millican, Malcolm Reynolds,
  et~al.
\newblock Flamingo: a visual language model for few-shot learning.
\newblock In \emph{NeurIPS}, 2022.

\bibitem[Bai et~al.(2023)Bai, Bai, Yang, Wang, Tan, Wang, Lin, Zhou, and
  Zhou]{bai2023qwen}
Jinze Bai, Shuai Bai, Shusheng Yang, Shijie Wang, Sinan Tan, Peng Wang, Junyang
  Lin, Chang Zhou, and Jingren Zhou.
\newblock {Qwen-VL}: A frontier large vision-language model with versatile
  abilities.
\newblock \emph{arXiv:2308.12966}, 2023.

\bibitem[Beyer et~al.(2024)Beyer, Steiner, Pinto, Kolesnikov, Wang, Salz,
  Neumann, Alabdulmohsin, Tschannen, Bugliarello, et~al.]{beyer2024paligemma}
Lucas Beyer, Andreas Steiner, Andr{\'e}~Susano Pinto, Alexander Kolesnikov,
  Xiao Wang, Daniel Salz, Maxim Neumann, Ibrahim Alabdulmohsin, Michael
  Tschannen, Emanuele Bugliarello, et~al.
\newblock Paligemma: A versatile 3b vlm for transfer.
\newblock \emph{arXiv:2407.07726}, 2024.

\bibitem[Chen et~al.(2023{\natexlab{a}})Chen, Zhu, Shen, Li, Liu, Zhang,
  Krishnamoorthi, Chandra, Xiong, and Elhoseiny]{chen2023minigpt}
Jun Chen, Deyao Zhu, Xiaoqian Shen, Xiang Li, Zechun Liu, Pengchuan Zhang,
  Raghuraman Krishnamoorthi, Vikas Chandra, Yunyang Xiong, and Mohamed
  Elhoseiny.
\newblock {MiniGPT-v2}: Large language model as a unified interface for
  vision-language multi-task learning.
\newblock \emph{arXiv:2310.09478}, 2023{\natexlab{a}}.

\bibitem[Chen et~al.(2023{\natexlab{b}})Chen, Li, Dong, Zhang, He, Wang, Zhao,
  and Lin]{chen2023sharegpt4v}
Lin Chen, Jisong Li, Xiaoyi Dong, Pan Zhang, Conghui He, Jiaqi Wang, Feng Zhao,
  and Dahua Lin.
\newblock {ShareGPT4V}: Improving large multi-modal models with better
  captions.
\newblock \emph{arXiv:2311.12793}, 2023{\natexlab{b}}.

\bibitem[Chen et~al.(2023{\natexlab{c}})Chen, Djolonga, Padlewski, Mustafa,
  Changpinyo, Wu, Ruiz, Goodman, Wang, Tay, et~al.]{chen2023palix}
Xi~Chen, Josip Djolonga, Piotr Padlewski, Basil Mustafa, Soravit Changpinyo,
  Jialin Wu, Carlos~Riquelme Ruiz, Sebastian Goodman, Xiao Wang, Yi~Tay, et~al.
\newblock {PaLI-X}: On scaling up a multilingual vision and language model.
\newblock \emph{arXiv:2305.18565}, 2023{\natexlab{c}}.

\bibitem[Chen et~al.(2023{\natexlab{d}})Chen, Wang, Beyer, Kolesnikov, Wu,
  Voigtlaender, Mustafa, Goodman, Alabdulmohsin, Padlewski,
  et~al.]{chen2023pali3}
Xi~Chen, Xiao Wang, Lucas Beyer, Alexander Kolesnikov, Jialin Wu, Paul
  Voigtlaender, Basil Mustafa, Sebastian Goodman, Ibrahim Alabdulmohsin, Piotr
  Padlewski, et~al.
\newblock {PaLI-3} vision language models: Smaller, faster, stronger.
\newblock \emph{arXiv:2310.09199}, 2023{\natexlab{d}}.

\bibitem[Chen et~al.(2023{\natexlab{e}})Chen, Wang, Changpinyo, Piergiovanni,
  Padlewski, Salz, Goodman, Grycner, Mustafa, Beyer, et~al.]{chen2023pali}
Xi~Chen, Xiao Wang, Soravit Changpinyo, AJ~Piergiovanni, Piotr Padlewski,
  Daniel Salz, Sebastian Goodman, Adam Grycner, Basil Mustafa, Lucas Beyer,
  et~al.
\newblock {PaLI}: A jointly-scaled multilingual language-image model.
\newblock In \emph{ICLR}, 2023{\natexlab{e}}.

\bibitem[Chen et~al.(2023{\natexlab{f}})Chen, Wu, Wang, Su, Chen, Xing, Zhong,
  Zhang, Zhu, Lu, Li, Luo, Lu, Qiao, and Dai]{chen2023internvl}
Zhe Chen, Jiannan Wu, Wenhai Wang, Weijie Su, Guo Chen, Sen Xing, Muyan Zhong,
  Qinglong Zhang, Xizhou Zhu, Lewei Lu, Bin Li, Ping Luo, Tong Lu, Yu~Qiao, and
  Jifeng Dai.
\newblock {InternVL}: Scaling up vision foundation models and aligning for
  generic visual-linguistic tasks.
\newblock \emph{arXiv:2312.14238}, 2023{\natexlab{f}}.

\bibitem[Chen et~al.(2024)Chen, Wang, Tian, Ye, Gao, Cui, Tong, Hu, Luo, Ma,
  et~al.]{chen2024far}
Zhe Chen, Weiyun Wang, Hao Tian, Shenglong Ye, Zhangwei Gao, Erfei Cui, Wenwen
  Tong, Kongzhi Hu, Jiapeng Luo, Zheng Ma, et~al.
\newblock How far are we to gpt-4v? closing the gap to commercial multimodal
  models with open-source suites.
\newblock \emph{arXiv:2404.16821}, 2024.

\bibitem[Cherti et~al.(2023)Cherti, Beaumont, Wightman, Wortsman, Ilharco,
  Gordon, Schuhmann, Schmidt, and Jitsev]{cherti2023reproducible}
Mehdi Cherti, Romain Beaumont, Ross Wightman, Mitchell Wortsman, Gabriel
  Ilharco, Cade Gordon, Christoph Schuhmann, Ludwig Schmidt, and Jenia Jitsev.
\newblock Reproducible scaling laws for contrastive language-image learning.
\newblock In \emph{CVPR}, 2023.

\bibitem[Chiang et~al.(2023)Chiang, Li, Lin, Sheng, Wu, Zhang, Zheng, Zhuang,
  Zhuang, Gonzalez, Stoica, and Xing]{vicuna2023}
Wei-Lin Chiang, Zhuohan Li, Zi~Lin, Ying Sheng, Zhanghao Wu, Hao Zhang, Lianmin
  Zheng, Siyuan Zhuang, Yonghao Zhuang, Joseph~E. Gonzalez, Ion Stoica, and
  Eric~P. Xing.
\newblock Vicuna: An open-source chatbot impressing gpt-4 with 90\%* chatgpt
  quality.
\newblock \url{https://lmsys.org/blog/2023-03-30-vicuna/}, March 2023.

\bibitem[Dai et~al.(2024)Dai, Li, Li, Tiong, Zhao, Wang, Li, Fung, and
  Hoi]{dai2024instructblip}
Wenliang Dai, Junnan Li, Dongxu Li, Anthony Meng~Huat Tiong, Junqi Zhao,
  Weisheng Wang, Boyang Li, Pascale~N Fung, and Steven Hoi.
\newblock {InstructBLIP}: Towards general-purpose vision-language models with
  instruction tuning.
\newblock In \emph{NeurIPS}, 2024.

\bibitem[Dong et~al.(2024)Dong, Zhang, Zang, Cao, Wang, Ouyang, Zhang, Duan,
  Zhang, Li, et~al.]{dong2024internlm2hd}
Xiaoyi Dong, Pan Zhang, Yuhang Zang, Yuhang Cao, Bin Wang, Linke Ouyang,
  Songyang Zhang, Haodong Duan, Wenwei Zhang, Yining Li, et~al.
\newblock {InternLM-XComposer2-4KHD}: A pioneering large vision-language model
  handling resolutions from 336 pixels to 4k hd.
\newblock \emph{arXiv:2404.06512}, 2024.

\bibitem[Driess et~al.(2023)Driess, Xia, Sajjadi, Lynch, Chowdhery, Ichter,
  Wahid, Tompson, Vuong, Yu, et~al.]{driess2023palm}
Danny Driess, Fei Xia, Mehdi~SM Sajjadi, Corey Lynch, Aakanksha Chowdhery,
  Brian Ichter, Ayzaan Wahid, Jonathan Tompson, Quan Vuong, Tianhe Yu, et~al.
\newblock {PaLM-E}: An embodied multimodal language model.
\newblock \emph{arXiv:2303.03378}, 2023.

\bibitem[Dubey et~al.(2024)Dubey, Jauhri, Pandey, Kadian, Al-Dahle, Letman,
  Mathur, Schelten, Yang, Fan, et~al.]{dubey2024llama}
Abhimanyu Dubey, Abhinav Jauhri, Abhinav Pandey, Abhishek Kadian, Ahmad
  Al-Dahle, Aiesha Letman, Akhil Mathur, Alan Schelten, Amy Yang, Angela Fan,
  et~al.
\newblock The {Llama} 3 herd of models.
\newblock \emph{arXiv:2407.21783}, 2024.

\bibitem[Fan et~al.(2024)Fan, Ji, Jiang, Li, Jin, Song, Wang, Hong, Chen,
  Zheng, et~al.]{fan2024mousi}
Xiaoran Fan, Tao Ji, Changhao Jiang, Shuo Li, Senjie Jin, Sirui Song, Junke
  Wang, Boyang Hong, Lu~Chen, Guodong Zheng, et~al.
\newblock {MouSi}: Poly-visual-expert vision-language models.
\newblock \emph{arXiv:2401.17221}, 2024.

\bibitem[Fang et~al.(2023{\natexlab{a}})Fang, Sun, Wang, Huang, Wang, and
  Cao]{fang2023eva02}
Yuxin Fang, Quan Sun, Xinggang Wang, Tiejun Huang, Xinlong Wang, and Yue Cao.
\newblock {EVA-02}: A visual representation for neon genesis.
\newblock \emph{arXiv:2303.11331}, 2023{\natexlab{a}}.

\bibitem[Fang et~al.(2023{\natexlab{b}})Fang, Wang, Xie, Sun, Wu, Wang, Huang,
  Wang, and Cao]{fang2023eva}
Yuxin Fang, Wen Wang, Binhui Xie, Quan Sun, Ledell Wu, Xinggang Wang, Tiejun
  Huang, Xinlong Wang, and Yue Cao.
\newblock {EVA}: Exploring the limits of masked visual representation learning
  at scale.
\newblock In \emph{CVPR}, 2023{\natexlab{b}}.

\bibitem[Fei et~al.(2024)Fei, Yao, Zhang, Liu, Zhang, and
  Chua]{fei2024multimodal}
Hao Fei, Yuan Yao, Zhuosheng Zhang, Fuxiao Liu, Ao~Zhang, and Tat-Seng Chua.
\newblock From multimodal llm to human-level ai: Modality, instruction,
  reasoning, efficiency and beyond.
\newblock In \emph{LREC-Coling Tutorials}, 2024.

\bibitem[Fu et~al.(2023)Fu, Chen, Shen, Qin, Zhang, Lin, Qiu, Lin, Yang, Zheng,
  et~al.]{fu2023mme}
Chaoyou Fu, Peixian Chen, Yunhang Shen, Yulei Qin, Mengdan Zhang, Xu~Lin,
  Zhenyu Qiu, Wei Lin, Jinrui Yang, Xiawu Zheng, et~al.
\newblock {MME}: A comprehensive evaluation benchmark for multimodal large
  language models.
\newblock \emph{arXiv:2306.13394}, 2023.

\bibitem[Gao et~al.(2023)Gao, Pi, Zhang, Ye, Zhong, Wang, Hong, Han, Xu, Li,
  and Kong]{geo170k}
Jiahui Gao, Renjie Pi, Jipeng Zhang, Jiacheng Ye, Wanjun Zhong, Yufei Wang,
  Lanqing Hong, Jianhua Han, Hang Xu, Zhenguo Li, and Lingpeng Kong.
\newblock G-llava: Solving geometric problem with multi-modal large language
  model.
\newblock \emph{arXiv:2312.11370}, 2023.

\bibitem[Goyal et~al.(2017)Goyal, Khot, Summers{-}Stay, Batra, and
  Parikh]{balanced_vqa_v2}
Yash Goyal, Tejas Khot, Douglas Summers{-}Stay, Dhruv Batra, and Devi Parikh.
\newblock Making the {V} in {VQA} matter: Elevating the role of image
  understanding in {V}isual {Q}uestion {A}nswering.
\newblock In \emph{CVPR}, 2017.

\bibitem[Gurari et~al.(2018)Gurari, Li, Stangl, Guo, Lin, Grauman, Luo, and
  Bigham]{2018vizwiz}
Danna Gurari, Qing Li, Abigale~J. Stangl, Anhong Guo, Chi Lin, Kristen Grauman,
  Jiebo Luo, and Jeffrey~P. Bigham.
\newblock Vizwiz grand challenge: Answering visual questions from blind people.
\newblock In \emph{CVPR}, 2018.

\bibitem[He et~al.(2024)He, Wei, Xie, and Tian]{he2024incorporating}
Xin He, Longhui Wei, Lingxi Xie, and Qi~Tian.
\newblock Incorporating visual experts to resolve the information loss in
  multimodal large language models.
\newblock \emph{arXiv:2401.03105}, 2024.

\bibitem[Hong et~al.(2024)Hong, Wang, Lv, Xu, Yu, Ji, Wang, Wang, Dong, Ding,
  et~al.]{hong2024cogagent}
Wenyi Hong, Weihan Wang, Qingsong Lv, Jiazheng Xu, Wenmeng Yu, Junhui Ji, Yan
  Wang, Zihan Wang, Yuxiao Dong, Ming Ding, et~al.
\newblock {CogAgent}: A visual language model for gui agents.
\newblock In \emph{CVPR}, 2024.

\bibitem[Hudson \& Manning(2019)Hudson and Manning]{hudson2019gqa}
Drew~A Hudson and Christopher~D Manning.
\newblock {GQA}: A new dataset for real-world visual reasoning and
  compositional question answering.
\newblock In \emph{CVPR}, 2019.

\bibitem[Ilharco et~al.(2021)Ilharco, Wortsman, Wightman, Gordon, Carlini,
  Taori, Dave, Shankar, Namkoong, Miller, Hajishirzi, Farhadi, and
  Schmidt]{ilharco_gabriel_2021_5143773}
Gabriel Ilharco, Mitchell Wortsman, Ross Wightman, Cade Gordon, Nicholas
  Carlini, Rohan Taori, Achal Dave, Vaishaal Shankar, Hongseok Namkoong, John
  Miller, Hannaneh Hajishirzi, Ali Farhadi, and Ludwig Schmidt.
\newblock Openclip, July 2021.
\newblock URL \url{https://doi.org/10.5281/zenodo.5143773}.
\newblock If you use this software, please cite it as below.

\bibitem[Kafle et~al.(2018)Kafle, Cohen, Price, and Kanan]{kafle2018dvqa}
Kushal Kafle, Scott Cohen, Brian Price, and Christopher Kanan.
\newblock {DVQA}: Understanding data visualizations via question answering.
\newblock In \emph{CVPR}, 2018.

\bibitem[Kar et~al.(2024)Kar, Tonioni, Poklukar, Kulshrestha, Zamir, and
  Tombari]{kar2024brave}
O{\u{g}}uzhan~Fatih Kar, Alessio Tonioni, Petra Poklukar, Achin Kulshrestha,
  Amir Zamir, and Federico Tombari.
\newblock {BRAVE}: Broadening the visual encoding of vision-language models.
\newblock \emph{arXiv:2404.07204}, 2024.

\bibitem[Karamcheti et~al.(2024)Karamcheti, Nair, Balakrishna, Liang, Kollar,
  and Sadigh]{karamcheti2024prismatic}
Siddharth Karamcheti, Suraj Nair, Ashwin Balakrishna, Percy Liang, Thomas
  Kollar, and Dorsa Sadigh.
\newblock Prismatic vlms: Investigating the design space of
  visually-conditioned language models.
\newblock \emph{arXiv:2402.07865}, 2024.

\bibitem[Kembhavi et~al.(2016{\natexlab{a}})Kembhavi, Salvato, Kolve, Seo,
  Hajishirzi, and Farhadi]{Kembhavi2016ADI}
Aniruddha Kembhavi, Michael Salvato, Eric Kolve, Minjoon Seo, Hannaneh
  Hajishirzi, and Ali Farhadi.
\newblock A diagram is worth a dozen images.
\newblock \emph{arXiv:1603.07396}, 2016{\natexlab{a}}.

\bibitem[Kembhavi et~al.(2016{\natexlab{b}})Kembhavi, Salvato, Kolve, Seo,
  Hajishirzi, and Farhadi]{ai2d}
Aniruddha Kembhavi, Mike Salvato, Eric Kolve, Min~Joon Seo, Hannaneh
  Hajishirzi, and Ali Farhadi.
\newblock A diagram is worth a dozen images.
\newblock In \emph{Computer Vision - {ECCV} 2016 - 14th European Conference,
  Amsterdam, The Netherlands, October 11-14, 2016, Proceedings, Part {IV}},
  volume 9908 of \emph{Lecture Notes in Computer Science}, pp.\  235--251,
  2016{\natexlab{b}}.

\bibitem[Kim et~al.(2022)Kim, Hong, Yim, Nam, Park, Yim, Hwang, Yun, Han, and
  Park]{kim2022donut}
Geewook Kim, Teakgyu Hong, Moonbin Yim, JeongYeon Nam, Jinyoung Park, Jinyeong
  Yim, Wonseok Hwang, Sangdoo Yun, Dongyoon Han, and Seunghyun Park.
\newblock {OCR-Free} document understanding transformer.
\newblock In \emph{ECCV}, 2022.

\bibitem[Kirillov et~al.(2023)Kirillov, Mintun, Ravi, Mao, Rolland, Gustafson,
  Xiao, Whitehead, Berg, Lo, et~al.]{kirillov2023segment}
Alexander Kirillov, Eric Mintun, Nikhila Ravi, Hanzi Mao, Chloe Rolland, Laura
  Gustafson, Tete Xiao, Spencer Whitehead, Alexander~C Berg, Wan-Yen Lo, et~al.
\newblock Segment anything.
\newblock In \emph{ICCV}, 2023.

\bibitem[Lee et~al.(2024)Lee, Park, Kim, and Ro]{lee2024moai}
Byung-Kwan Lee, Beomchan Park, Chae~Won Kim, and Yong~Man Ro.
\newblock {MoAI}: Mixture of all intelligence for large language and vision
  models.
\newblock \emph{arXiv:2403.07508}, 2024.

\bibitem[Lee et~al.(2023)Lee, Joshi, Turc, Hu, Liu, Eisenschlos, Khandelwal,
  Shaw, Chang, and Toutanova]{lee2023pix2struct}
Kenton Lee, Mandar Joshi, Iulia~Raluca Turc, Hexiang Hu, Fangyu Liu,
  Julian~Martin Eisenschlos, Urvashi Khandelwal, Peter Shaw, Ming-Wei Chang,
  and Kristina Toutanova.
\newblock {Pix2Struct}: Screenshot parsing as pretraining for visual language
  understanding.
\newblock In \emph{ICML}, 2023.

\bibitem[Li et~al.(2023{\natexlab{a}})Li, Zhang, Chen, Wang, Pu, Yang, Li, and
  Liu]{li2023mimic}
Bo~Li, Yuanhan Zhang, Liangyu Chen, Jinghao Wang, Fanyi Pu, Jingkang Yang,
  Chunyuan Li, and Ziwei Liu.
\newblock {MIMIC-IT}: Multi-modal in-context instruction tuning.
\newblock \emph{arXiv:2306.05425}, 2023{\natexlab{a}}.

\bibitem[Li et~al.(2023{\natexlab{b}})Li, Zhang, Chen, Wang, Yang, and
  Liu]{li2023otter}
Bo~Li, Yuanhan Zhang, Liangyu Chen, Jinghao Wang, Jingkang Yang, and Ziwei Liu.
\newblock Otter: A multi-modal model with in-context instruction tuning.
\newblock \emph{arXiv:2305.03726}, 2023{\natexlab{b}}.

\bibitem[Li et~al.(2023{\natexlab{c}})Li, Wang, Wang, Ge, Ge, and
  Shan]{li2023seed}
Bohao Li, Rui Wang, Guangzhi Wang, Yuying Ge, Yixiao Ge, and Ying Shan.
\newblock {Seed-Bench}: Benchmarking multimodal llms with generative
  comprehension.
\newblock \emph{arXiv:2307.16125}, 2023{\natexlab{c}}.

\bibitem[Li et~al.(2024{\natexlab{a}})Li, Gan, Yang, Yang, Li, Wang, Gao,
  et~al.]{li2024multimodal}
Chunyuan Li, Zhe Gan, Zhengyuan Yang, Jianwei Yang, Linjie Li, Lijuan Wang,
  Jianfeng Gao, et~al.
\newblock Multimodal foundation models: From specialists to general-purpose
  assistants.
\newblock \emph{Foundations and Trends{\textregistered} in Computer Graphics
  and Vision}, 2024{\natexlab{a}}.

\bibitem[Li et~al.(2022)Li, Li, Xiong, and Hoi]{li2022blip}
Junnan Li, Dongxu Li, Caiming Xiong, and Steven Hoi.
\newblock {BLIP}: Bootstrapping language-image pre-training for unified
  vision-language understanding and generation.
\newblock In \emph{ICML}, 2022.

\bibitem[Li et~al.(2023{\natexlab{d}})Li, Li, Savarese, and Hoi]{li2023blip}
Junnan Li, Dongxu Li, Silvio Savarese, and Steven Hoi.
\newblock {BLIP-2}: Bootstrapping language-image pre-training with frozen image
  encoders and large language models.
\newblock In \emph{ICML}, 2023{\natexlab{d}}.

\bibitem[Li et~al.(2024{\natexlab{b}})Li, Zhang, Wang, Zhong, Chen, Chu, Liu,
  and Jia]{li2024mini}
Yanwei Li, Yuechen Zhang, Chengyao Wang, Zhisheng Zhong, Yixin Chen, Ruihang
  Chu, Shaoteng Liu, and Jiaya Jia.
\newblock {Mini-Gemini}: Mining the potential of multi-modality vision language
  models.
\newblock \emph{arXiv:2403.18814}, 2024{\natexlab{b}}.

\bibitem[Li et~al.(2023{\natexlab{e}})Li, Du, Zhou, Wang, Zhao, and
  Wen]{li2023evaluating}
Yifan Li, Yifan Du, Kun Zhou, Jinpeng Wang, Wayne~Xin Zhao, and Ji-Rong Wen.
\newblock Evaluating object hallucination in large vision-language models.
\newblock \emph{arXiv:2305.10355}, 2023{\natexlab{e}}.

\bibitem[Li et~al.(2024{\natexlab{c}})Li, Yang, Liu, Ma, Zhang, Yang, Sun, Liu,
  and Bai]{li2024monkey}
Zhang Li, Biao Yang, Qiang Liu, Zhiyin Ma, Shuo Zhang, Jingxu Yang, Yabo Sun,
  Yuliang Liu, and Xiang Bai.
\newblock Monkey: Image resolution and text label are important things for
  large multi-modal models.
\newblock In \emph{CVPR}, 2024{\natexlab{c}}.

\bibitem[Lin et~al.(2023{\natexlab{a}})Lin, Yin, Ping, Lu, Molchanov, Tao, Mao,
  Kautz, Shoeybi, and Han]{lin2023vila}
Ji~Lin, Hongxu Yin, Wei Ping, Yao Lu, Pavlo Molchanov, Andrew Tao, Huizi Mao,
  Jan Kautz, Mohammad Shoeybi, and Song Han.
\newblock {VILA}: On pre-training for visual language models.
\newblock \emph{arXiv:2312.07533}, 2023{\natexlab{a}}.

\bibitem[Lin et~al.(2023{\natexlab{b}})Lin, Liu, Zhang, Gao, Qiu, Xiao, Qiu,
  Lin, Shao, Chen, et~al.]{lin2023sphinx}
Ziyi Lin, Chris Liu, Renrui Zhang, Peng Gao, Longtian Qiu, Han Xiao, Han Qiu,
  Chen Lin, Wenqi Shao, Keqin Chen, et~al.
\newblock {SPHINX}: The joint mixing of weights, tasks, and visual embeddings
  for multi-modal large language models.
\newblock \emph{arXiv:2311.07575}, 2023{\natexlab{b}}.

\bibitem[Liu et~al.(2023{\natexlab{a}})Liu, Guan, Li, Chen, Yacoob, Manocha,
  and Zhou]{liu2023hallusionbench}
Fuxiao Liu, Tianrui Guan, Zongxia Li, Lichang Chen, Yaser Yacoob, Dinesh
  Manocha, and Tianyi Zhou.
\newblock {HallusionBench}: You see what you think? or you think what you see?
  an image-context reasoning benchmark challenging for gpt-4v (ision),
  llava-1.5, and other multi-modality models.
\newblock \emph{arXiv:2310.14566}, 2023{\natexlab{a}}.

\bibitem[Liu et~al.(2023{\natexlab{b}})Liu, Lin, Li, Wang, Yacoob, and
  Wang]{liu2023aligning}
Fuxiao Liu, Kevin Lin, Linjie Li, Jianfeng Wang, Yaser Yacoob, and Lijuan Wang.
\newblock Aligning large multi-modal model with robust instruction tuning.
\newblock \emph{arXiv:2306.14565}, 2023{\natexlab{b}}.

\bibitem[Liu et~al.(2023{\natexlab{c}})Liu, Li, Li, and Lee]{liu2023improved}
Haotian Liu, Chunyuan Li, Yuheng Li, and Yong~Jae Lee.
\newblock Improved baselines with visual instruction tuning.
\newblock \emph{arXiv:2310.03744}, 2023{\natexlab{c}}.

\bibitem[Liu et~al.(2023{\natexlab{d}})Liu, Li, Wu, and Lee]{liu2023visual}
Haotian Liu, Chunyuan Li, Qingyang Wu, and Yong~Jae Lee.
\newblock Visual instruction tuning.
\newblock In \emph{NeurIPS}, 2023{\natexlab{d}}.

\bibitem[Liu et~al.(2024{\natexlab{a}})Liu, Li, Li, Li, Zhang, Shen, and
  Lee]{liu2024llavanext}
Haotian Liu, Chunyuan Li, Yuheng Li, Bo~Li, Yuanhan Zhang, Sheng Shen, and
  Yong~Jae Lee.
\newblock {LLaVA-NeXT}: Improved reasoning, ocr, and world knowledge.
\newblock \url{https://llava-vl.github.io/blog/2024-01-30-llava-next/}, January
  2024{\natexlab{a}}.

\bibitem[Liu et~al.(2024{\natexlab{b}})Liu, Fan, Johns, Yu, Xiao, and
  Anandkumar]{liu2024prismer}
Shikun Liu, Linxi Fan, Edward Johns, Zhiding Yu, Chaowei Xiao, and Anima
  Anandkumar.
\newblock Prismer: A vision-language model with an ensemble of experts.
\newblock \emph{TMLR}, 2024{\natexlab{b}}.

\bibitem[Liu et~al.(2023{\natexlab{e}})Liu, Duan, Zhang, Li, Zhang, Zhao, Yuan,
  Wang, He, Liu, Chen, and Lin]{MMBench}
Yuan Liu, Haodong Duan, Yuanhan Zhang, Bo~Li, Songyang Zhang, Wangbo Zhao, Yike
  Yuan, Jiaqi Wang, Conghui He, Ziwei Liu, Kai Chen, and Dahua Lin.
\newblock {MMBench}: Is your multi-modal model an all-around player?
\newblock \emph{arXiv:2307.06281}, 2023{\natexlab{e}}.

\bibitem[Liu et~al.(2023{\natexlab{f}})Liu, Li, Yang, Li, Yin, lin Liu, Jin,
  and Bai]{liu2023hidden}
Yuliang Liu, Zhang Li, Biao Yang, Chunyuan Li, Xucheng Yin, Cheng lin Liu,
  Lianwen Jin, and Xiang Bai.
\newblock On the hidden mystery of ocr in large multimodal models.
\newblock \emph{arXiv:2305.07895}, 2023{\natexlab{f}}.

\bibitem[Liu et~al.(2022)Liu, Mao, Wu, Feichtenhofer, Darrell, and
  Xie]{liu2022convnet}
Zhuang Liu, Hanzi Mao, Chao-Yuan Wu, Christoph Feichtenhofer, Trevor Darrell,
  and Saining Xie.
\newblock A convnet for the 2020s.
\newblock In \emph{CVPR}, 2022.

\bibitem[Lu et~al.(2024)Lu, Bansal, Xia, Liu, Li, Hajishirzi, Cheng, Chang,
  Galley, and Gao]{lu2024mathvista}
Pan Lu, Hritik Bansal, Tony Xia, Jiacheng Liu, Chunyuan Li, Hannaneh
  Hajishirzi, Hao Cheng, Kai-Wei Chang, Michel Galley, and Jianfeng Gao.
\newblock {MathVista}: Evaluating mathematical reasoning of foundation models
  in visual contexts.
\newblock In \emph{ICLR}, 2024.

\bibitem[Luo et~al.(2024)Luo, Zhou, Zhang, Zheng, Sun, and Ji]{luo2024feast}
Gen Luo, Yiyi Zhou, Yuxin Zhang, Xiawu Zheng, Xiaoshuai Sun, and Rongrong Ji.
\newblock Feast your eyes: Mixture-of-resolution adaptation for multimodal
  large language models.
\newblock \emph{arXiv:2403.03003}, 2024.

\bibitem[Masry et~al.(2022)Masry, Long, Tan, Joty, and Hoque]{masry2022chartqa}
Ahmed Masry, Do~Xuan Long, Jia~Qing Tan, Shafiq Joty, and Enamul Hoque.
\newblock {ChartQA}: A benchmark for question answering about charts with
  visual and logical reasoning.
\newblock \emph{arXiv:2203.10244}, 2022.

\bibitem[Mathew et~al.(2021)Mathew, Karatzas, and Jawahar]{mathew2021docvqa}
Minesh Mathew, Dimosthenis Karatzas, and C.~V. Jawahar.
\newblock {DocVQA}: A dataset for vqa on document images.
\newblock In \emph{WACV}, 2021.

\bibitem[Oquab et~al.(2023)Oquab, Darcet, Moutakanni, Vo, Szafraniec, Khalidov,
  Fernandez, Haziza, Massa, El-Nouby, et~al.]{oquab2023dinov2}
Maxime Oquab, Timoth{\'e}e Darcet, Th{\'e}o Moutakanni, Huy Vo, Marc
  Szafraniec, Vasil Khalidov, Pierre Fernandez, Daniel Haziza, Francisco Massa,
  Alaaeldin El-Nouby, et~al.
\newblock {DINOv2}: Learning robust visual features without supervision.
\newblock \emph{arXiv:2304.07193}, 2023.

\bibitem[Radford et~al.(2021)Radford, Kim, Hallacy, Ramesh, Goh, Agarwal,
  Sastry, Askell, Mishkin, Clark, et~al.]{radford2021learning}
Alec Radford, Jong~Wook Kim, Chris Hallacy, Aditya Ramesh, Gabriel Goh,
  Sandhini Agarwal, Girish Sastry, Amanda Askell, Pamela Mishkin, Jack Clark,
  et~al.
\newblock Learning transferable visual models from natural language
  supervision.
\newblock In \emph{ICML}, 2021.

\bibitem[Ranzinger et~al.(2024)Ranzinger, Heinrich, Kautz, and
  Molchanov]{radio}
Mike Ranzinger, Greg Heinrich, Jan Kautz, and Pavlo Molchanov.
\newblock {AM-RADIO}: Agglomerative vision foundation model reduce all domains
  into one.
\newblock In \emph{CVPR}, 2024.

\bibitem[Saikh et~al.(2022)Saikh, Ghosal, Mittal, Ekbal, and
  Bhattacharyya]{saikh2022scienceqa}
Tanik Saikh, Tirthankar Ghosal, Amish Mittal, Asif Ekbal, and Pushpak
  Bhattacharyya.
\newblock {ScienceQA}: A novel resource for question answering on scholarly
  articles.
\newblock \emph{International Journal on Digital Libraries}, 2022.

\bibitem[Schuhmann et~al.(2022)Schuhmann, Beaumont, Vencu, Gordon, Wightman,
  Cherti, Coombes, Katta, Mullis, Wortsman, Schramowski, Kundurthy, Crowson,
  Schmidt, Kaczmarczyk, and Jitsev]{schuhmann2022laionb}
Christoph Schuhmann, Romain Beaumont, Richard Vencu, Cade~W Gordon, Ross
  Wightman, Mehdi Cherti, Theo Coombes, Aarush Katta, Clayton Mullis, Mitchell
  Wortsman, Patrick Schramowski, Srivatsa~R Kundurthy, Katherine Crowson,
  Ludwig Schmidt, Robert Kaczmarczyk, and Jenia Jitsev.
\newblock {LAION-5B}: An open large-scale dataset for training next generation
  image-text models.
\newblock In \emph{NeurIPS Datasets and Benchmarks Track}, 2022.
\newblock URL \url{https://openreview.net/forum?id=M3Y74vmsMcY}.

\bibitem[Shi et~al.(2024)Shi, Wu, Mao, Wang, and Darrell]{shi2024need}
Baifeng Shi, Ziyang Wu, Maolin Mao, Xin Wang, and Trevor Darrell.
\newblock When do we not need larger vision models?
\newblock \emph{arXiv:2403.13043}, 2024.

\bibitem[Shi et~al.(2016)Shi, Caballero, Huszar, Totz, Aitken, Bishop,
  Rueckert, and Wang]{pixelshffule}
Wenzhe Shi, Jose Caballero, Ferenc Huszar, Johannes Totz, Andrew~P. Aitken, Rob
  Bishop, Daniel Rueckert, and Zehan Wang.
\newblock Real-time single image and video super-resolution using an efficient
  sub-pixel convolutional neural network.
\newblock In \emph{CVPR}, 2016.

\bibitem[Singh et~al.(2019)Singh, Natarajan, Shah, Jiang, Chen, Batra, Parikh,
  and Rohrbach]{singh2019towards}
Amanpreet Singh, Vivek Natarajan, Meet Shah, Yu~Jiang, Xinlei Chen, Dhruv
  Batra, Devi Parikh, and Marcus Rohrbach.
\newblock Towards {VQA} models that can read.
\newblock In \emph{CVPR}, 2019.

\bibitem[Sun et~al.(2023)Sun, Fang, Wu, Wang, and Cao]{sun2023eva}
Quan Sun, Yuxin Fang, Ledell Wu, Xinlong Wang, and Yue Cao.
\newblock {EVA-CLIP}: Improved training techniques for clip at scale.
\newblock \emph{arXiv:2303.15389}, 2023.

\bibitem[Team et~al.(2023)Team, Anil, Borgeaud, Wu, Alayrac, Yu, Soricut,
  Schalkwyk, Dai, Hauth, et~al.]{team2023gemini}
Gemini Team, Rohan Anil, Sebastian Borgeaud, Yonghui Wu, Jean-Baptiste Alayrac,
  Jiahui Yu, Radu Soricut, Johan Schalkwyk, Andrew~M Dai, Anja Hauth, et~al.
\newblock Gemini: a family of highly capable multimodal models.
\newblock \emph{arXiv:2312.11805}, 2023.

\bibitem[Teknium(2023)]{OpenHermes2.5}
Teknium.
\newblock {OpenHermes 2.5}: An open dataset of synthetic data for generalist
  {LLM} assistants.
\newblock \url{https://huggingface.co/datasets/teknium/OpenHermes-2.5}, 2023.

\bibitem[Tong et~al.(2024)Tong, Brown, Wu, Woo, Middepogu, Akula, Yang, Yang,
  Iyer, Pan, Wang, Fergus, LeCun, and Xie]{tong2024cambrian}
Shengbang Tong, Ellis Brown, Penghao Wu, Sanghyun Woo, Manoj Middepogu,
  Sai~Charitha Akula, Jihan Yang, Shusheng Yang, Adithya Iyer, Xichen Pan,
  Austin Wang, Rob Fergus, Yann LeCun, and Saining Xie.
\newblock Cambrian-1: A fully open, vision-centric exploration of multimodal
  llms.
\newblock \emph{arXiv:2406.16860}, 2024.

\bibitem[Wang et~al.(2023{\natexlab{a}})Wang, Meng, Weng, He, Wu, and
  Jiang]{wang2023instruct4v}
Junke Wang, Lingchen Meng, Zejia Weng, Bo~He, Zuxuan Wu, and Yu-Gang Jiang.
\newblock To see is to believe: Prompting gpt-4v for better visual instruction
  tuning.
\newblock \emph{arXiv:2311.07574}, 2023{\natexlab{a}}.

\bibitem[Wang et~al.(2023{\natexlab{b}})Wang, Lv, Yu, Hong, Qi, Wang, Ji, Yang,
  Zhao, Song, et~al.]{wang2023cogvlm}
Weihan Wang, Qingsong Lv, Wenmeng Yu, Wenyi Hong, Ji~Qi, Yan Wang, Junhui Ji,
  Zhuoyi Yang, Lei Zhao, Xixuan Song, et~al.
\newblock {CogVLM}: Visual expert for pretrained language models.
\newblock \emph{arXiv:2311.03079}, 2023{\natexlab{b}}.

\bibitem[Wu et~al.(2024)Wu, Xian, Guan, Liang, Chakraborty, Liu, Sadler,
  Manocha, and Bedi]{wu2024safety}
Xiyang Wu, Ruiqi Xian, Tianrui Guan, Jing Liang, Souradip Chakraborty, Fuxiao
  Liu, Brian Sadler, Dinesh Manocha, and Amrit~Singh Bedi.
\newblock On the safety concerns of deploying llms/vlms in robotics:
  Highlighting the risks and vulnerabilities.
\newblock \emph{arXiv:2402.10340}, 2024.

\bibitem[xAI(2024)]{RWQA}
xAI.
\newblock {Grok-1.5 Vision Preview}.
\newblock \url{https://x.ai/blog/grok-1.5v}, 2024.

\bibitem[{xAI}(2024)]{xai2024grok}
{xAI}.
\newblock Grok-1.5v.
\newblock \url{https://x.ai/blog/grok-1.5v}, 2024.
\newblock Accessed: 2024-09-28.

\bibitem[Xu et~al.(2024)Xu, Yao, Guo, Cui, Ni, Ge, Chua, Liu, and
  Huang]{xu2024llava-uhd}
Ruyi Xu, Yuan Yao, Zonghao Guo, Junbo Cui, Zanlin Ni, Chunjiang Ge, Tat-Seng
  Chua, Zhiyuan Liu, and Gao Huang.
\newblock {LLaVA-UHD}: an lmm perceiving any aspect ratio and high-resolution
  images.
\newblock \emph{arXiv:2403.11703}, 2024.

\bibitem[Yin et~al.(2024)Yin, Fu, Zhao, Li, Sun, Xu, and Chen]{yin2024survey}
Shukang Yin, Chaoyou Fu, Sirui Zhao, Ke~Li, Xing Sun, Tong Xu, and Enhong Chen.
\newblock A survey on multimodal large language models.
\newblock \emph{IEEE Trans. PAMI}, 2024.

\bibitem[Yue et~al.(2024)Yue, Ni, Zhang, Zheng, Liu, Zhang, Stevens, Jiang,
  Ren, Sun, et~al.]{yue2024mmmu}
Xiang Yue, Yuansheng Ni, Kai Zhang, Tianyu Zheng, Ruoqi Liu, Ge~Zhang, Samuel
  Stevens, Dongfu Jiang, Weiming Ren, Yuxuan Sun, et~al.
\newblock {MMMU}: A massive multi-discipline multimodal understanding and
  reasoning benchmark for expert {AGI}.
\newblock In \emph{CVPR}, 2024.

\bibitem[Zeng et~al.(2023)Zeng, Zhang, Zheng, Xia, Wei, Wei, Zhang, and
  Kong]{zeng2023matters}
Yan Zeng, Hanbo Zhang, Jiani Zheng, Jiangnan Xia, Guoqiang Wei, Yang Wei,
  Yuchen Zhang, and Tao Kong.
\newblock What matters in training a gpt4-style language model with multimodal
  inputs?
\newblock \emph{arXiv:2307.02469}, 2023.

\bibitem[Zhai et~al.(2023)Zhai, Mustafa, Kolesnikov, and Beyer]{siglip}
Xiaohua Zhai, Basil Mustafa, Alexander Kolesnikov, and Lucas Beyer.
\newblock Sigmoid loss for language image pre-training.
\newblock In \emph{ICCV}, 2023.

\bibitem[Zhang et~al.(2023)Zhang, Zhang, Gu, Zhou, Lipka, Yang, and
  Sun]{zhang2023llavar}
Yanzhe Zhang, Ruiyi Zhang, Jiuxiang Gu, Yufan Zhou, Nedim Lipka, Diyi Yang, and
  Tong Sun.
\newblock {LLaVAR}: Enhanced visual instruction tuning for text-rich image
  understanding.
\newblock \emph{arXiv:2306.17107}, 2023.

\bibitem[Zhu et~al.(2023)Zhu, Chen, Shen, Li, and Elhoseiny]{zhu2023minigpt}
Deyao Zhu, Jun Chen, Xiaoqian Shen, Xiang Li, and Mohamed Elhoseiny.
\newblock {MiniGPT-4}: Enhancing vision-language understanding with advanced
  large language models.
\newblock \emph{arXiv:2304.10592}, 2023.

\bibitem[Zhu et~al.(2021)Zhu, Su, Lu, Li, Wang, and Dai]{deformable-detr}
Xizhou Zhu, Weijie Su, Lewei Lu, Bin Li, Xiaogang Wang, and Jifeng Dai.
\newblock {Deformable DETR}: Deformable transformers for end-to-end object
  detection.
\newblock In \emph{ICLR}, 2021.

\bibitem[Zhu et~al.(2016)Zhu, Groth, Bernstein, and Fei-Fei]{zhu2016cvpr}
Yuke Zhu, Oliver Groth, Michael Bernstein, and Li~Fei-Fei.
\newblock {Visual7W: Grounded Question Answering in Images}.
\newblock In \emph{CVPR}, 2016.

\bibitem[Zong et~al.(2024)Zong, Ma, Shen, Song, Shao, Jiang, Li, and
  Liu]{zong2024mova}
Zhuofan Zong, Bingqi Ma, Dazhong Shen, Guanglu Song, Hao Shao, Dongzhi Jiang,
  Hongsheng Li, and Yu~Liu.
\newblock {MoVA}: Adapting mixture of vision experts to multimodal context.
\newblock \emph{arXiv:2404.13046}, 2024.

\end{thebibliography}
\bibliographystyle{iclr2025}

\clearpage
\appendix
\section{Appendix}
\label{sec:appendix}

\subsection{Benchmark details}
\label{sec:appendix_bench_detail}

In this section, we provide the additional benchmark details for the tables in Section~\ref{sec:preliminary}. The detailed comparison of different adaptation methods for the CLIP encoder are shown in Table~\ref{tab:resolution-complete}.
Table~\ref{tab:different-vision-experts-complete} shows the comparison between different vision encoders on all the adopted benchmarks.
Table~\ref{tab:fusion-complete} list the detailed results on the vision encoder fusion methods.
Table~\ref{tab:vision-encoder-combination-details} shows the comparison between different vision encoder combinations on each benchmark.

\begin{table}[h]
    \caption{\textbf{Comparison of different high-resolution adaption methods to strengthen CLIP model (336x336).} RWQA denotes the RealworldQA~\citep{RWQA}.
    }
    \label{tab:resolution-complete}
    \small
    \centering
    \renewcommand{\arraystretch}{1.2} 
    \addtolength{\tabcolsep}{0pt}
    \resizebox{\textwidth}{!}{
    \begin{tabular}{ccccccccccccccccc}
    \toprule[1pt]
    Vision Encoder & Unfreeze & Res & GQA & VizWiz & MME & SEED & OCR & DocVQA & ChartQA & AI2D & POPE & RWQA & SQA & Avg.  \\
    \midrule
    \textit{Original}       & \ding{55}     & 336 & 63.2 & 55.1 & 1574 & 70.2 & 354 & 59.6 & 42.1 & 71.1 & 86.7 & 58.0        & 72.5    & 616.5 
    \\
    \textit{Original }      & \ding{51}      & 336 & 60.9 & 54.0 & 1501 & 61.8 & 305 & 56.4 & 24.5 & 69.3 & 80.9 & 53.6        & 72.0    & 562.6 
    \\
    \midrule
    \textit{Interpolate}    & \ding{55}     & 448 & 62.7 & 53.8  & 1470 & 69.1 & 299 & 58.4 & 37.7 & 71.2 & 85.7 & 55.3        & 69.6    & 589.7 
    \\
    \textit{Interpolate}   & \ding{51}      & 448 & 65.6 & 57.8  & 1534 & 73.7 & 526 & 65.0 & 61.1 & 73.7 & 87.3 & 57.7        & 71.5    & 670.5 
    \\
    \textit{Interpolate}    & \ding{51}      & 672 & 64.9 & 55.7 & 1503 & 72.4 & 509 & 64.6 & 62.0 & 72.2 & 87.1 & 57.4        & 71.2    & 674.2 
    \\
    \textit{Tiled-input}    & \ding{51}      & 672 & 63.0 & 54.9 & 1529 & 72.5 & 435 & 64.9 & 65.7 & 71.5 & 87.6 & 57.0        & 71.4    & 673.9 
    \\
    \midrule
    \midrule
    \textit{InternVL}       & \ding{55}     & 448 & 63.6 & 56.9 & 1537 & 71.7 & 529 & 65.0 & 58.6 & 72.9 & 87.4 & 59.2        & 70.2    & 661.9 
    \\
    \textit{InternVL}      & \ding{51}      & 448 & 65.6 & 57.8 & 1534 & 73.7 & 526 & 65.0 & 61.1 & 73.7 & 87.3 & 58.8        & 71.5    & 671.5 
    \\
    \bottomrule[1pt]
    \end{tabular}
    }
\end{table}

\begin{table}[h]
    \caption{\textbf{Comparison between different vision experts as the MLLM encoders.}}
    \label{tab:different-vision-experts-complete}
    \small
    \centering
    \renewcommand{\arraystretch}{1.2} 
    \addtolength{\tabcolsep}{-2pt}
    \resizebox{\textwidth}{!}{
    \begin{tabular}{cccccccccccccccc}
    \toprule[1pt]
    Category & Vision Encoder  & Unfreeze &  Res & GQA & VizWiz & MME & SEED & OCR & DocVQA & ChartQA & AI2D & POPE & RWQA & SQA & Avg.\\
    \midrule
    \multirow{2}{*}{\textit{VL Alignment}} & \multirow{2}{*}{\textit{ConvNeXt}} & \ding{55} & 1024 & 63.3 & 53.5 & 1526 & 70.6 & 404 & 70.4 & 60.8 & 71.6 & 87.5 & 57.1 & 68.6 & 635.0 \\
    & & \ding{51} & 1024 & 63.3 & 54.4 & 1510 & 72.9 & 518 & 77.9 & 67.0 & 72.1 & 88.1 & 58.8 & 68.6 & 659.7 \\
    \midrule
    \multirow{2}{*}{\textit{Segmentation}} & \multirow{2}{*}{\textit{SAM}} & \ding{55} & 1024 & 57.3 & 49.0 & 1216 & 56.9 & 38 & 20.1 & 17.4 & 69.2 & 84.3 & 49.2 & 66.8 & 471.1 \\
    & & \ding{51} & 1024 & 60.2 & 51.5 & 1291 & 65.9 & 35 & 21.2 & 17.8 & 70.7 & 86.4 & 54.1 & 65.7 & 494.7 \\
    \midrule
    \multirow{2}{*}{\textit{Object Detection}} & \multirow{2}{*}{\textit{EVA-02}} & \ding{55} & 1024 & 63.1 & 51.1 & 1359 & 69.2 & 123 & 25.6 & 25.2 & 71.2 & 88.5 & 57.9 & 66.1 & 523.5 \\
    & & \ding{51} & 1024 & 64.3 & 55.5 & 1449 & 72.7 & 358 & 57.1 & 57.5 & 72.2 & 88.3 & 59.6 & 67.7 & 614.4 \\
    \midrule
    \multirow{2}{*}{\textit{Text Recognition}} & \multirow{2}{*}{\textit{Pix2Struct}} & \ding{55} & 1024 & 53.1 & 48.1 & 1296 & 53.4 & 460 & 71.0 & 61.0 & 69.6 & 79.2 & 46.7 & 65.5 & 578.7 \\
    & & \ding{51} & 1024 & 54.9 & 47.3 & 1262 & 55.1 & 472 & 72.5 & 62.0 & 68.7 & 80.0 & 49.3 & 66.6 & 584.8 \\
    \midrule
    \multirow{2}{*}{\textit{Self-Supervised}} & \multirow{2}{*}{\textit{DINOv2}} & \ding{55} & 448 & 62.4 & 53.1 & 1438 & 67.4 & 41 & 20.2 & 17.3 & 70.7 & 85.3 & 53.3 & 67.1 & 503.1 \\
    & & \ding{51} & 448 & 64.2 & 55.5 & 1466 & 71.8 & 45 & 20.4 & 17.5 & 71.4 & 87.4 & 57.3 & 67.8 & 518.0 \\
    \bottomrule[1pt]
    \end{tabular}
    }
\end{table}

\begin{table}[h]
    \caption{\textbf{Comparison of different fusion methods for different vision experts.} ``\#Tokens(V)'' denotes the number of visual tokens.}
    \label{tab:fusion-complete}
    \small
    \centering
    \renewcommand{\arraystretch}{1.2} 
    \addtolength{\tabcolsep}{-1pt}
    \resizebox{\textwidth}{!}{
    \begin{tabular}{c|ccccccccccccccc}
    \toprule[1pt]
    Vision Encoders & Fusion & \#Tokens(V) & GQA & VizWiz & MME & SEED & OCR & DocVQA & ChartQA & AI2D & POPE & RWQA & SQA & Avg. \\
    \midrule
    \multirow{5}{*}{\textit{CLIP} + \textit{ConvNeXt}} & \textit{Seq. Append} & 2048 & 64.8 & 54.5 & 1563 & 73.4 & 532 & 77.7 & 67.6 & 72.4 & 87.9 & 61.2 & 68.8 & 690.5 
    \\
    & \textit{Channel Concat.} & 1024 & 63.2 & 48.0 & 1497 & 73.5 & 551 & 77.7 & 67.0 & 72.4 & 88.3 & 59.1 & 70.7 & 681.5 
    \\
    & \textit{LLaVA-HR} & 1024 & 64.5 & 57.2& 1538 & 72.0 & 498 & 74.5 & 63.8 & 72.3 & 87.7 & 59.2 & 68.7 & 678.7 
    \\
    & \textit{Mini-Gemini} & 1024 & 65.3 & 56.9 & 1548 & 72.9 & 478 & 68.3 & 63.2 & 71.5 & 87.3 & 59.7 & 69.4 & 672.5 
    \\
    & \textit{Deformable Attn.} & 1024 & 64.0 & 57.3 & 1504 & 72.7 & 463 & 69.5 & 64.4 & 73.3 & 87.4 & 62.8 & 68.9 & 674.3 
    \\
    \midrule
    \multirow{2}{*}{\begin{tabular}[c]{@{}c@{}}\textit{CLIP} + \textit{ConvNeXt} \\ + \textit{SAM}\end{tabular}} & \textit{Seq. Append} & 3072 & 64.3 & 53.6 & 1539 & 73.2 & 525 & 77.9 & 67.0 & 72.3 & 87.4 & 60.1 & 69.5 & 686.2 
    \\
    & \textit{Channel Concat.} & 1024 & 63.3 & 55.9 & 1528 & 73.3 & 545 & 78.9 & 67.2 & 72.3 & 88.4 & 59.2 & 70.0 & 690.4 
    \\
    \bottomrule[1pt]
    \end{tabular}
}
\end{table}

\begin{table}[t]
    \caption{\textbf{Detailed comparison on vision encoder combinations.}}
    \label{tab:vision-encoder-combination-details}
    \small
    \centering
    \renewcommand{\arraystretch}{1.2} 
    \addtolength{\tabcolsep}{-2pt}
    \resizebox{\textwidth}{!}{
    \begin{tabular}{c|lcccccccccccc}
    \toprule[1pt]
    \#Encoder          & Encoder Combination         & GQA & VizWiz  & MME    & SEED & OCR   & DocVQA & ChartQA & AI2D & POPE & RWQA & SQA  & Avg.  \\
    \midrule
    2                  & CL + CN                     & 63.2 & 48.0 & 1497.0 & 73.5 & 551.0 & 77.7   & 67.0    &                      72.4 & 88.3 & 59.1 & 70.7 & 681.5 \\
    \midrule
    \multirow{4}{*}{3} & CL + CN + DI                & 63.3 & 55.9 & 1528.0 & 73.3 & 545.0 & 78.9   & 67.2    &                      72.3 & 88.4 & 59.2 & 70.0 & 690.4 \\
                       & CL + CN + SA                & 64.6 & 55.3 & 1504.0 & 73.3 & 526.0 & 75.7   & 64.9    & 72.1 & 88.3 & 61.1 & 70.9 & 685.4 \\
                       & CL + CN + PS                & 63.2 & 51.4 & 1497.0 & 73.3 & 550.0 & 78.5   & 65.9    & 73.1 & 87.7 & 60.3 & 70.5 & 685.1 \\
                       & CL + CN + EV                & 63.2 & 51.7 & 1565.0 & 73.9 & 538.0 & 77.7   & 67.8    & 73.6 & 89.0 & 61.4 & 69.4 & 690.7 \\
                       \midrule
                       & CL + CN + EV + DI           & 63.6 & 54.9 & 1512.0 & 73.8 & 547.0 & 77.0   & 66.7    & 73.1 & 88.9 & 60.4 & 69.7 & 689.4 \\
                       & CL + CN + EV + SA           & 64.3 & 57.7 & 1533.0 & 73.7 & 521.0 & 75.2   & 65.3    & 72.2 & 88.5 & 61.1 & 70.2 & 688.0 \\
                       & CL + CN + EV + PS           & 64.8 & 56.5 & 1561.0 & 73.4 & 540.0 & 78.8   & 67.5    & 72.2 & 88.4 & 59.9 & 70.5 & 694.6 \\
                       \midrule
    \multirow{2}{*}{5} & CL + CN + EV + PS + SA      & 64.7 & 59.1 & 1528.0 & 73.9 & 529.0 & 78.6   & 67.8    &                      72.9 & 88.8 & 62.2 & 69.7 & 697.1 \\
                       & CL + CN + EV + PS + DI      & 64.7 & 54.1 & 1506.0 & 73.7 & 541.0 & 75.1   & 64.9    & 72.7 & 88.3 & 60.0 & 70.3 & 684.7 \\
                       \midrule
    6                  & CL + CN + EV + PS + SA + DI & 63.8 & 57.8 & 1512.0 & 73.5 & 525.0 & 75.1   & 65.8    & 71.8 & 88.4 & 61.4 & 69.9 & 686.8 \\
    \bottomrule[1pt]
    \end{tabular}
    }
\end{table}

\begin{table}[!t]
    \caption{\textbf{Comparison between different training strategies.} `1 epoch'' means we train \textit{Eagle} for 1 epoch in the supervised fine-tuning stage. `unlock*'' means we unlock vision encoders in the pre-training stage.}
    \label{tab:training_stratgy_more}
    \small
    \centering
    \renewcommand{\arraystretch}{1.2} 
    \addtolength{\tabcolsep}{-4pt}
    \resizebox{\textwidth}{!}{
    \begin{tabular}{cccccccccccccccccc}
    \toprule
    Config & Prealign & Pretrain & Finetune &  \rotatebox{90}{GQA}   & \rotatebox{90}{MME}  &  \rotatebox{90}{OCR}   & \rotatebox{90}{SciQA}  & \rotatebox{90}{POPE}   & \rotatebox{90}{DocVQA} & \rotatebox{90}{ChartQA} & \rotatebox{90}{SEED} & \rotatebox{90}{Vizwiz}  & \rotatebox{90}{AI2D}  & \rotatebox{90}{RWQA} & Avg.  \\ \midrule
    
1 epoch       & \ding{55}     & llava595k & Eagle1.8M      & 64.7& 	1528	& 52.9	& 69.7	& 88.8& 	78.6& 	67.7	& 73.9& 	59.1& 	72.8& 	62.2& 697.1
\\
2 epoch       & \ding{55}     & llava595k & Eagle1.8M      & 65.4&	1539&	51.4	&70.3	&87.9&	79.8&	67.9&	73.8	&58.5	&73.5&	62.7&698.3
\\
1 epoch, unlock*       & \ding{55}     & llava595k & Eagle1.8M      & 64.1&	1541	&54.4	&71.5&	88.5&	79.1&	68.5&	74.0&	56.6&	72.2&	61.9& 698.0
\\
1 epoch, unlock*       & \ding{55}     & llava595k+Eagle1.8M & Eagle1.8M      & 65.3	 &1545 &	54.8 &	70.5 &	88.5	 &78.8 &	68.4	 &73.5 &	57.5 &	72.0 &	62.9 & \textbf{699.5}
\\
\hline
1 epoch       & Eagle1.8M     & llava595k & Eagle1.8M      & 64.5&	1579	&55.4&	71.6	&88.7&	79.3	&68.9&	73.9	&58.9	&73.9	&63.4&706.6
\\
1 epoch, unlock*       & Eagle1.8M     & llava595k & Eagle1.8M      & 65.1&	1561	&56.0	&71.3&	88.5&	79.3&	69.5	&74.2&	59&	73.5&	63.4&707.1
\\
1 epoch, unlock*       & llava595k+Eagle1.8M     & llava595k & Eagle1.8M      & 64.5	&1570&	56.0	&71.7&	89&	79.7&	68.9&	74.0&	58.5	&74.0&	63.7&707.8
\\
1 epoch, unlock*       & llava595k+Eagle1.8M     & llava595k+Eagle1.8M & Eagle1.8M      & 64.6&	1582&	56.6&	71.9&	88.7&	79.8&	69.3	&73.7&	58.7&	73.5&	63.9&\textbf{708.9}
\\ \bottomrule
    \end{tabular}
    }
\vspace{-0.2in}
\end{table}

\clearpage
\subsection{Additional qualitative results}

\begin{figure*}[h]
    \centering
      \includegraphics[width=\textwidth]{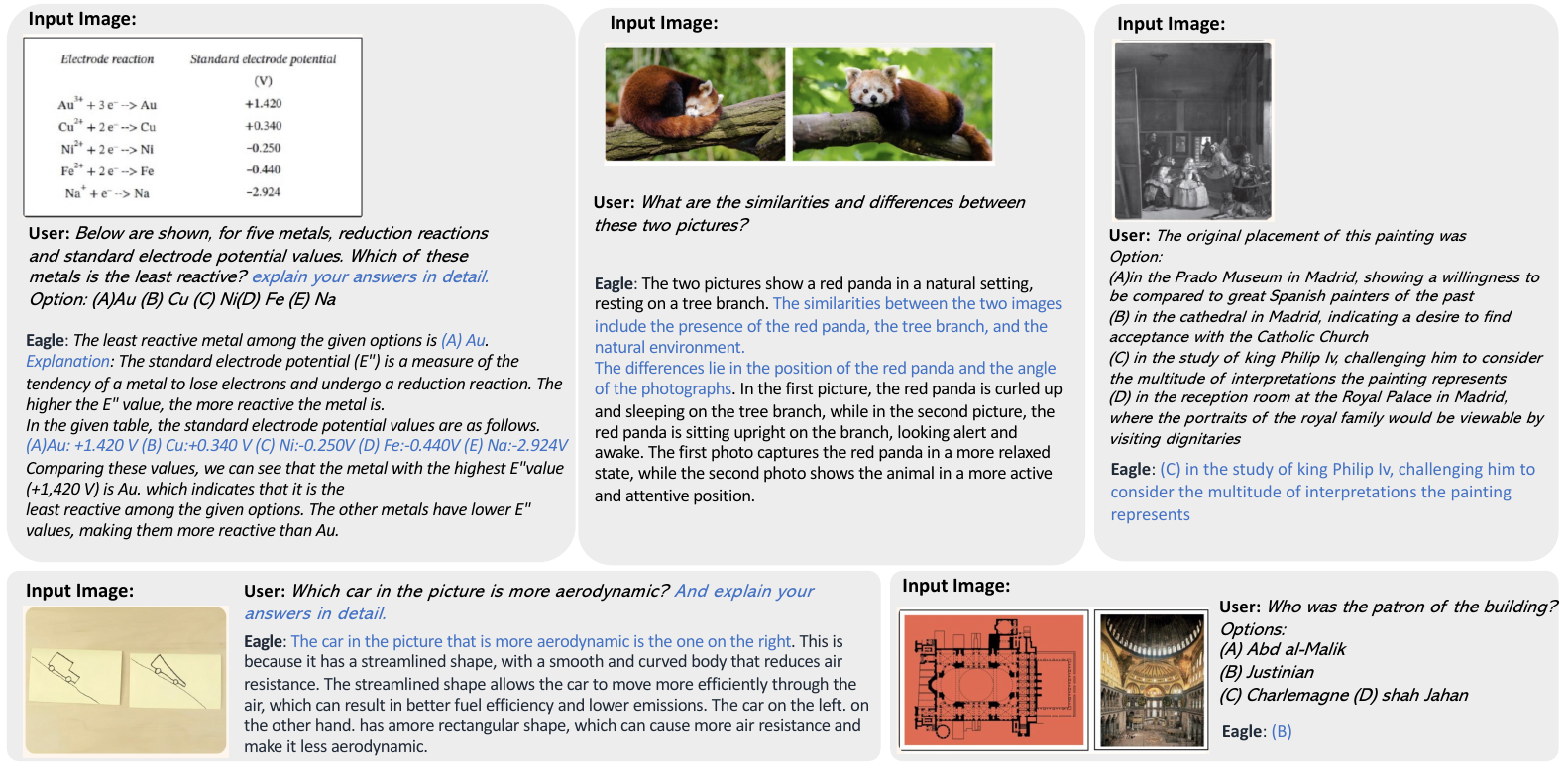}
    \caption{
    \textbf{Qualitative results of Eagle} on VQA tasks requiring college-level knowledge or deliberate reasoning.
    }
    \label{fig:visual_knowledge}
\end{figure*}

\begin{figure*}[h]
    \centering
      \includegraphics[width=\textwidth]{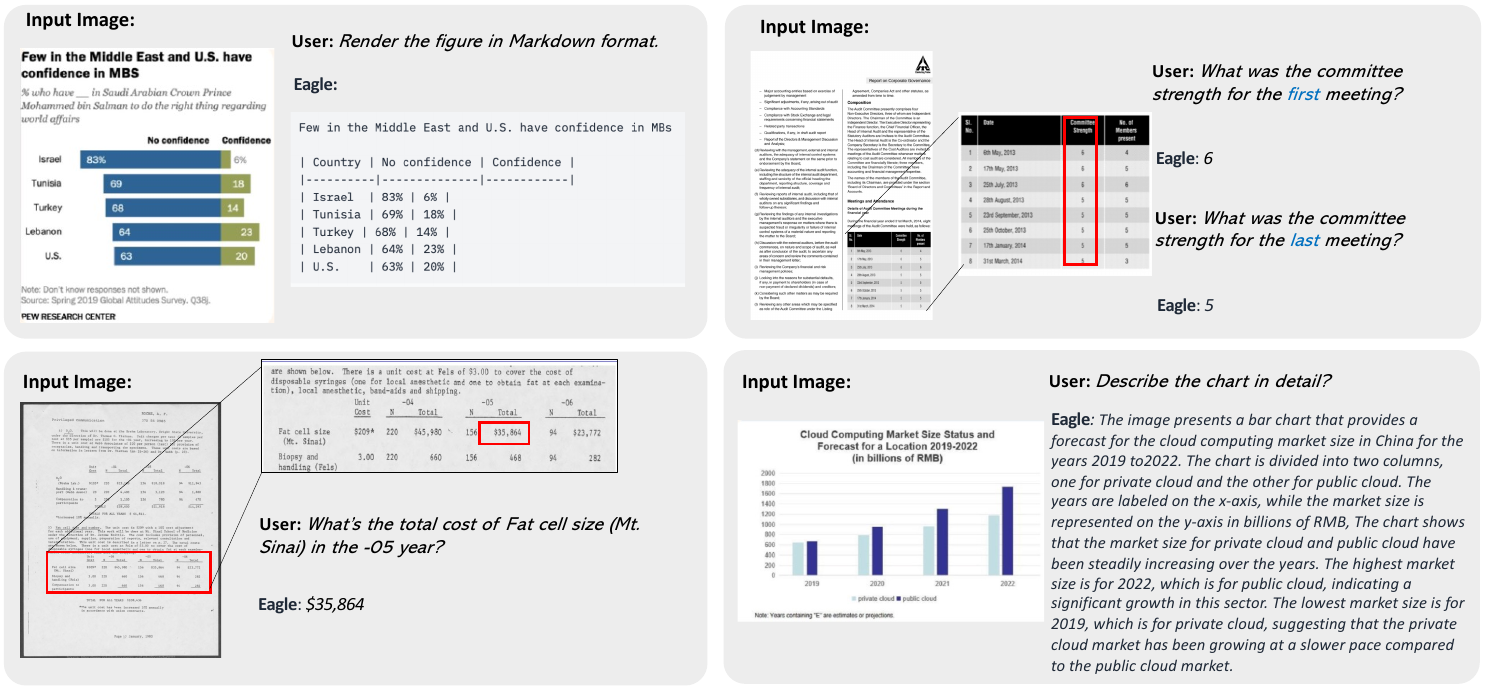}
    \caption{
    \textbf{Qualitative samples on OCR and document understanding tasks}. Eagle is able to extract useful information from small text.
    }
    \label{fig:qualitative-ocr}
\end{figure*}

Fig.~\ref{fig:visual_knowledge} demonstrates that our model can perform reasoning based on the knowledge to answer some difficult questions. Fig.~\ref{fig:qualitative-ocr} shows some examples of OCR and document understanding cases.

\end{document}